\documentclass[10pt,journal,compsoc]{IEEEtran}
\ifCLASSOPTIONcompsoc
  \usepackage[nocompress]{cite}
\else
  \usepackage{cite}
\fi
\ifCLASSINFOpdf
\else
\fi

\usepackage{amsmath,amsfonts}
\usepackage{algorithmic}
\usepackage{algorithm}
\usepackage{array}
\usepackage[caption=false,font=normalsize,labelfont=sf,textfont=sf]{subfig}
\usepackage{textcomp}
\usepackage{stfloats}
\usepackage{url}
\usepackage{verbatim}
\usepackage{graphicx}
\usepackage{cite}

\usepackage{hyperref}
\usepackage{booktabs}
\usepackage{graphicx}
\usepackage{multirow}
\usepackage{wrapfig}
\usepackage{bm}
\usepackage[table,xcdraw]{xcolor}
\usepackage{threeparttable}
\usepackage{siunitx}

\usepackage{amsthm}
\usepackage{amsmath}
\usepackage{array}

\newtheorem{remark}{Remark}
\usepackage[switch]{lineno}

\begin{document}
%
\title{On Exploring Node-feature and Graph-structure Diversities for Node Drop Graph Pooling}
%
%
%
%

\author{Chuang~Liu,
        Yibing~Zhan,
        Baosheng~Yu,
        Liu~Liu,
        Bo~Du,~\IEEEmembership{Senior Member,~IEEE},
        Wenbin~Hu,
        and~Tongliang~Liu,~\IEEEmembership{Senior Member,~IEEE} 
\IEEEcompsocitemizethanks{\IEEEcompsocthanksitem C. Liu is with the Computer School, Wuhan University, Hubei, China. 
E-mail:  chuangliu@whu.edu.cn.
\IEEEcompsocthanksitem Y. Zhan is currently an algorithm scientist at the JD Explore Academy, Beijing, China. 
E-mail: zhanyibing@jd.com.
\IEEEcompsocthanksitem B. Yu  is with the School of Computer Science, in the Faculty of Engineering, at The University of Sydney, 6 Cleveland St, Darlington, NSW 2008, Australia. 
E-mail: baosheng.yu@sydney.edu.au.
\IEEEcompsocthanksitem L. Liu  is with the School of Computer Science, in the Faculty of Engineering, at The University of Sydney, 6 Cleveland St, Darlington, NSW 2008, Australia. 
E-mail: liu.liu1@sydney.edu.au.
\IEEEcompsocthanksitem B. Du is with the Computer School, Wuhan University, Hubei, China. 
E-mail: dubo@whu.edu.cn.
\IEEEcompsocthanksitem W. Hu is with the Computer School, Wuhan University, Hubei, China, Corresponding Author \{W. Hu\}, 
E-mail: hwb@whu.edu.cn.
\IEEEcompsocthanksitem T. Liu is with the Trustworthy Machine Learning Lab, School of Computer Science, University of Sydney, Sydney, Building J12, 1 Cleveland St, Camperdown NSW 2006, Australia. 
E-mail: tongliang.liu@sydney.edu.au.}
\thanks{Manuscript received April 19, 2005; revised August 26, 2015.}}

%
%

\markboth{Journal of \LaTeX\ Class Files,~Vol.~14, No.~8, August~2015}%
{Shell \MakeLowercase{\textit{et al.}}: Bare Demo of IEEEtran.cls for Computer Society Journals}
%



\IEEEtitleabstractindextext{%
\begin{abstract}
A pooling operation is essential for effective graph-level representation learning, where the node drop pooling has become one mainstream graph pooling technology. However, current node drop pooling methods usually keep the top-k nodes according to their significance scores, which ignore the graph diversity in terms of the node features and the graph structures, thus resulting in suboptimal graph-level representations. To address the aforementioned issue, we propose a novel plug-and-play score scheme and refer to it as MID, which consists of a \textbf{M}ultidimensional score space with two operations, \textit{i.e.}, fl\textbf{I}pscore and \textbf{D}ropscore. Specifically, the multidimensional score space depicts the significance of nodes through multiple criteria; the flipscore encourages the maintenance of dissimilar node features; and the dropscore forces the model to notice diverse graph structures instead of being stuck in significant local structures. To evaluate the effectiveness of our proposed MID, we perform extensive experiments by applying it to a wide variety of recent node drop pooling methods, including TopKPool, SAGPool, GSAPool, and ASAP. Specifically, the proposed MID can efficiently and consistently achieve about 2.8\% average improvements over the above four methods on seventeen real-world graph classification datasets, including four social datasets (IMDB-BINARY, IMDB-MULTI, REDDIT-BINARY, and COLLAB), and thirteen biochemical datasets (D\&D, PROTEINS, NCI1, MUTAG, PTC-MR, NCI109, ENZYMES, MUTAGENICITY, FRANKENSTEIN, HIV, BBBP, TOXCAST, and TOX21). Code is available at~\url{https://github.com/whuchuang/mid}.
\end{abstract}
}

\maketitle

\IEEEdisplaynontitleabstractindextext

%
\IEEEpeerreviewmaketitle

\IEEEraisesectionheading{\section{Introduction}\label{sec:introduction}}

%
%
%
%

\IEEEPARstart{G}{raph} Neural Networks (GNNs) have achieved remarkable performance on a wide variety of graph-based tasks, including node classification~\cite{tkde-node-1,tkde-node-2,tkde-node-3}, link prediction~\cite{link1, link2, tkde-link-1}, and graph classification~\cite{fair-graph-classification, graphCL, gcc, transformer, sugar, tkde-graph-1, tkde-graph-3}. In node classification and link prediction, GNNs propagate information between nodes via graph convolutions, whereas in graph classification, information of all nodes is integrated together to generate graph-level representations through graph pooling. Early adopted global pooling, including average pooling and max pooling, ignores the node correlations, limiting the overall performance~\cite{duvenaud,set2set, sortpool}. Later, graph pooling utilizes hierarchical architectures to model the node correlations~\cite{eigenpool, second-order, TAPool} and can be roughly classified into two types: node clustering pooling and node drop pooling. Node clustering pooling requires clustering nodes into new nodes, which is time-and space-consuming~\cite{diffpool, mincut, structpool}. In contrast, node drop pooling only preserves the representative nodes by calculating the significance of nodes, and is more efficient and more fit for large-scale graphs~\cite{graph-u-net, sagpool, gsapool,ipool}.


\begin{figure*}[!t] 
\setlength{\abovecaptionskip}{-0.1cm}   
\begin{center}
\includegraphics[width=1.0\linewidth]{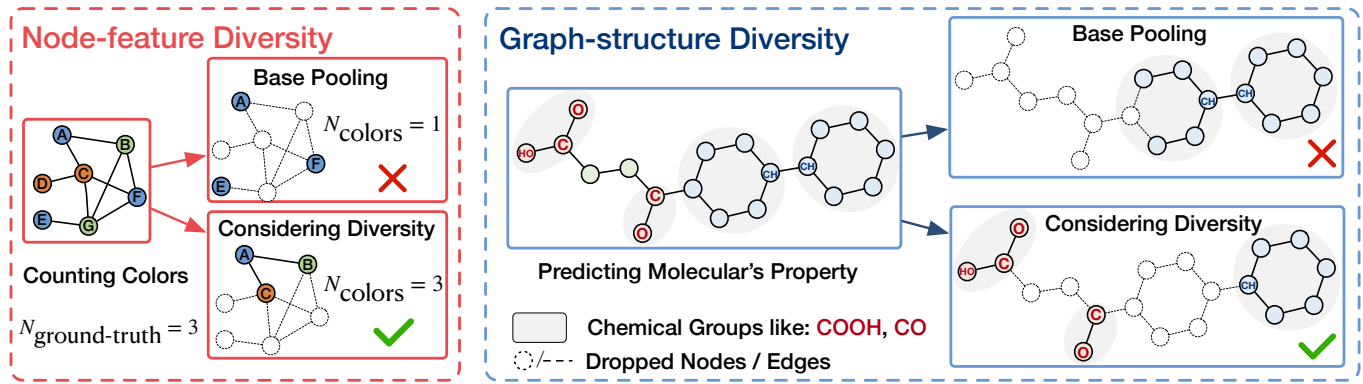}
\end{center}
\caption{Illustration of the node-feature and the graph-structure diversities. The left part  ({\color[HTML]{FE0000} in red block}) presents the task of counting colors in a graph, where each color represents a unique discrete feature.  The right part ({\color[HTML]{3166FF} in blue block})  presents the task of predicting the property of molecular, where chemical groups like OH (hydroxyl group), COOH (carboxyl group), CO (ketone group) as well as benzene rings have a great impact on the  property of molecules such as solubility. }
\label{fig:diversity}
\end{figure*}


Though the efficient and effective, current node drop pooling methods still obtain suboptimal graph-level representations as they ignore the diversities in graphs, including the node-feature diversity and the graph-structure diversity, by selecting only the top-k significant nodes. \textbf{Node-feature Diversity.} Analytically, current drop node pooling methods tend to highlight nodes with similar features, and thus perform less satisfactorily when multiple dissimilar nodes contribute to the graph-level representations. We leverage the insights of~\cite{understand-attention} and design a simple graph reasoning task that allows us to demonstrate the significance of the node-feature diversity. As shown in the left part of Fig.~\ref{fig:diversity}, the task is to count the number of colors in a graph, where a color is a unique discrete feature. Obviously, base pooling models would fail to predict the number of colors since  the reserved nodes are in the same color with the same score. Therefore, it is worthwhile to consider the node-feature diversity as the node feature plays an important role in the task. \textbf{Graph-structure Diversity.} Furthermore, connected nodes would share similar information through GNNs~\cite{birds,sgcn}. Current node drop pooling methods tend to be stuck into the nodes of local structures without consideration of the rest of representative graph structures. The experimental and theoretical verifications of the above intuition are presented in detail in Section~\ref{sec:method}. To demonstrate the significance of the graph-structure diversity, we introduce a task to predict the property of molecular. As shown in the right part of Fig.~\ref{fig:diversity}, base pooling models could only capture a single structure (benzene rings), while neglecting other chemical groups like OH (hydroxyl group), COOH (carboxyl group), CO (ketone group), which are all indispensable to the prediction of property of molecules, such as solubility~\cite{mem-pool,moleculenet}. Therefore, it is worthwhile to consider the graph-structure diversity as the node feature plays an important role in the task. 

To explore the diversity, we propose a novel plug-and-play scheme, termed MID, using a \textbf{M}ultidimensional score space with fl\textbf{I}pscore and \textbf{D}ropscore operations. Analytically, the multidimensional score space, with each dimension corresponding to one view of the nodes, depicts the significance of nodes by using vectors instead of scalars, which enables a comprehensive description of the attributes of nodes; flipscore reverses the negative confidence scores as positive ones, encouraging the node drop pooling models to highlight nodes with dissimilar features; dropscore randomly drops several nodes when selecting the top-k scores, probably deleting the nodes in local structures and thus forcing the node drop pooling to notice diverse graph structures.

We conduct extensive experiments for MID by applying MID to four typical node drop pooling methods (SAGPool~\cite{sagpool}, TopKPool~\cite{graph-u-net}, GSAPool~\cite{gsapool}, and ASAP~\cite{asap}). The results on seventeen real-world graph classification datasets, which vary in content domains and dataset sizes, demonstrate the ability and generalizability of MID. Specifically, MID consistently brings about 2.8\% improvements in average across all backbone models and datasets. Moreover, ablation experiments further demonstrate the contribution of individual components of MID to the exploration of graph diversities.
The main contributions of this paper are summarized as follows: 
\begin{itemize}
  \item We propose a new modularized framework for node drop pooling and analyze the distinctions and similarities among twelve typical models under the framework. 
  \item We propose a novel play-and-plug scheme, MID, which comprises a multidimensional score space, flipscore operation, and dropscore operation. Our MID could be applied in most node drop pooling models and boost their performance with relatively low computational cost. 
  \item We conduct extensive experiments for four typical drop node pooling methods with and without MID on the graph classiﬁcation task across seventeen real-world datasets as well as two synthetic datasets. The experimental results comprehensively demonstrate the effectiveness of MID.
\end{itemize}

\section{Related Work}
\label{graph-pooling}

\subsection{Graph Convolution Networks}

Recently, numerous researches have been proposed based on Graph Convolution Networks (GCNs), which generalize the convolution operation to graph data. The basic idea behind such methods as Graph Convolutional Network (GCN)~\cite{gcn}, GraphSAGE~\cite{graphsage},  Graph Attention Network (GAT)~\cite{gat}, and Graph Isomorphism Network (GIN)~\cite{gin}, is updating the embedding of each node with messages from its neighbor nodes. The methods mentioned above have achieved attractive performance on node classification and link prediction tasks. However, with the main aim of generating accurate node representations, these methods fail to obtain the entire graph representations in the absence of pooling operations.

\subsection{Graph Pooling} 
With a crucial role in representing the entire graph, graph pooling could be roughly divided into global pooling and hierarchical pooling. Global pooling performs global sum/average/max-pooling~\cite{duvenaud} or more sophisticated operations~\cite{set2set, sortpool, deepsets, msnapool, degree-pool, gin, tkde-graph-2} on all node features to obtain the graph-level representations, which causes information loss since they ignore the structures of graphs. Later, hierarchical pooling models are proposed  in consideration of the graph structure, which can be classified into node clustering pooling, node drop pooling, and other pooling models.  \textbf{1) Node clustering pooling} considers the graph pooling problem as a node clustering problem to map the nodes into a set of clusters~\cite{diffpool, lap-pool, NMFPool, mincut, eigenpool, structpool, HIBpool, gmt}, which  suffer from a limitation of time and storage complexity caused by the computation of the dense soft-assignment matrix.  Besides, as discussed in the recent works~\cite{rethink-pooling, understanding-pooling}, clustering-enforcing regularization is usually innocuous. \textbf{2) Node drop pooling} uses learnable scoring functions to delete nodes with comparatively  lower significance scores~\cite{topkpool, graph-u-net, sagpool, attpool, asap, hgp-sl, vip-pool, rep-pool, gsapool, path-pooling, NDPool, uniform-pooling, lookhops, cgi-pool, TAPool, ipool, MVpool, commpool}. Though more efficient and more applicable to  large-scale networks than node clustering pooling, node drop pooling suffers from an  inevitable information loss (including the diversity). \textbf{3) Other pooling.} Apart from 1) and 2), there exist some other graph pooling models. EdgePool~\cite{edgepool} contracts the high score edge between two nodes; HaarPool~\cite{haar-pooling, haarnet} compresses the node features in the Haar wavelet domain; MemPool~\cite{mem-pool} proposes an efficient memory layer to jointly learn node representations and coarsen the graph; SOPool~\cite{second-order} introduces the second-order statistics into graph coarsening; MuchPool~\cite{much-pool} combines the node clustering pooling and the node drop pooling to capture different characteristics of a graph; and PAS~\cite{pooling-search} proposes to search for adaptive pooling architectures by neural architecture search.

\section{Method}
\label{sec:method}

\begin{table}[!t]
    
    \caption{Commonly used notations and their descriptions.}
    \scriptsize
    \centering
    \renewcommand\arraystretch{1.3}
    \begin{tabular}{p{22pt}<{\centering}p{79pt}p{22pt}<{\centering}p{79pt}}
        \toprule
        \textbf{Notations} & \quad \textbf{Descriptions} & \textbf{Notations} & \quad\textbf{Descriptions} \\
        \midrule
        $\mathcal{G}$ & A graph & $\mathcal{E}$ & The set of edges in a graph  \\
        $\mathcal{V}$ & The set of nodes in a graph & $\boldsymbol{A}/ \boldsymbol{A}^{l} $ & The adjacency matrix (in layer $l$)  \\ $\bm{X} / \bm{X}^{l}$ & The matrix of node features (in layer $l$) & 
        $\boldsymbol{S}/ \boldsymbol{S}^{l} $ & The matrix of scores generated by the pooling models (in layer $l$)  \\
        $\mathrm{idx} / \mathrm{idx}^{l}$     & The preserving node indexes (in layer $l$) &
        $n$ & Number of nodes in a graph \\
        $c$       & The dimension of a node feature  &
        $k$ & The pooling ratio \\
        $p_s$       & The score dropping rate  &
        $h$ & The dimension of a score for a node \\
        $\odot$       & The broadcasted elementwise product & $\|$       & Vector concatenation  \\
        
        $\left\| \cdot \right\|_{1}$ & Manhattan norm ($L_1$ norm) &
        $\left\|\cdot\right\|_{2}$& Euclidean Distance ($L_2$ norm)
        \\ $\lceil{\cdot} \rceil$ & The operation of rounding up & $\left| \cdot \right|$ & Absolute value
        \\

                \bottomrule
    \end{tabular}
    \label{tab:notion}
\end{table}

In this section, we first propose a framework of current node drop pooling methods. Then, we present the details of our MID and explain how MID advances node drop pooling.

\textbf{Notation.}  Let $\mathcal{G}=(\mathcal{V}, \mathcal{E})$ denote a graph with node set $\mathcal{V}$ and edge set $\mathcal{E}$. The node features are denoted as $\boldsymbol{X} \in \mathbb{R}^{ n \times c}$, where  $ n$ is the number of nodes and $ c$ is the dimension of  node features. The adjacency matrix is defined as  $\boldsymbol{A} \in \{0, 1\}^{ n \times n}$. Notations are introduced in Table~\ref{tab:notion}.

\subsection{Proposed Node Drop Pooling Framework}

\begin{figure}[!t] 
\setlength{\abovecaptionskip}{-0.1cm}   
\begin{center}
\includegraphics[width=1.0\linewidth]{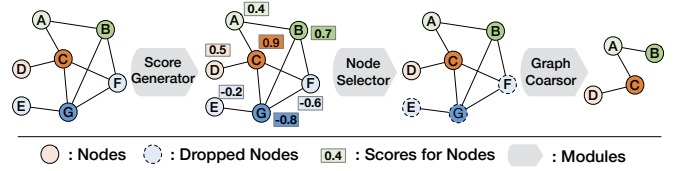}
\end{center}
\caption{Our modularized framework for node drop pooling.
}
\label{fig:framework}
\end{figure}

\begin{figure*}[!t] 
\setlength{\abovecaptionskip}{-0.1cm}   
\begin{center}
\includegraphics[width=0.75\linewidth]{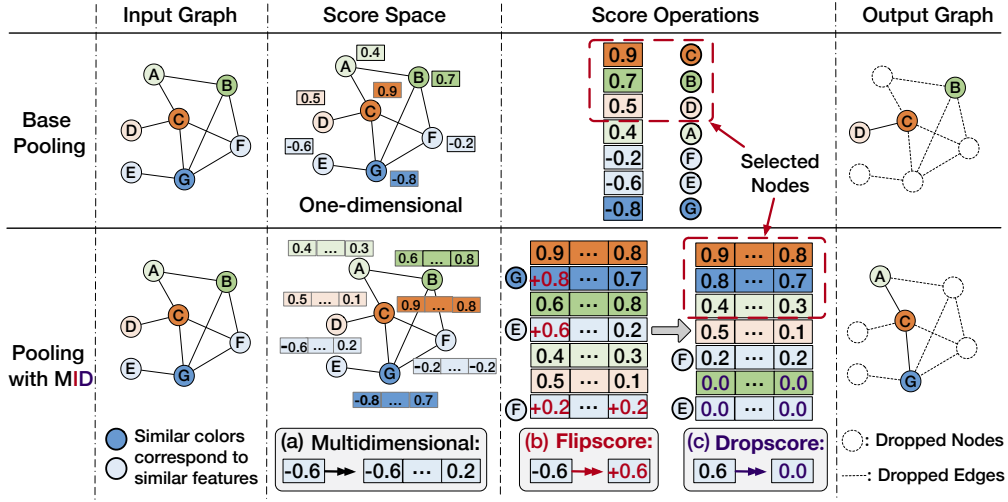}
\end{center}
\caption{Architecture of our proposed method. The order of flipscore and dropscore operations does not influence the pooling results.}
\label{fig:model}
\end{figure*}

\begin{table*}[!t]
\setlength{\belowcaptionskip}{0.3cm}   
\renewcommand\arraystretch{1.5} 
\centering

\caption{Summary of node drop pooling models in our framework\tnote{1}.}
\label{tab:schemes}
\begin{threeparttable}[b]
\scriptsize
\resizebox{\textwidth}{!}{%
\begin{tabular}{@{}llll@{}}
\toprule
\multicolumn{1}{l}{\normalsize \textbf{Models} }                                            & \multicolumn{1}{l}{\normalsize  \textbf{Score Generator} }                                                                                         & \multicolumn{1}{l}{\normalsize  \textbf{Node Selector} }                                                                                                                  & \multicolumn{1}{l}{\normalsize  \textbf{Graph Coarsor} }                                                                                                                                                                                                                                                                                                                                                                     \\ \midrule
\begin{tabular}[c]{@{}l@{}}TopKPool~\cite{topkpool,graph-u-net}\end{tabular} & $\boldsymbol{S}=\boldsymbol{X} \mathbf{p} /\left\|\mathbf{p}\right\|_{2}$                                                                                 & $\mathrm{idx}=\text{TOP}_{k}(\boldsymbol{S})$                                                                                              & $\begin{aligned}\boldsymbol{X}^{\prime}=\boldsymbol{X}_{\mathrm{idx}} \odot \sigma(\boldsymbol{S}_{\mathrm{idx}}); \quad \boldsymbol{A}^{\prime}=\boldsymbol{A}_{\mathrm{idx}, \mathrm{idx}}\end{aligned}$                                                                                                                                                                                                                                                                       \\ 
\begin{tabular}[c]{@{}l@{}}SAGPool~\cite{sagpool}\end{tabular}      & $\boldsymbol{S}=\sigma\left(\tilde{\boldsymbol{D}}^{-\frac{1}{2}} \tilde{\boldsymbol{A}} \tilde{\boldsymbol{D}}^{-\frac{1}{2}} \boldsymbol{X}\boldsymbol{W}\right)$                         & $\mathrm{idx}=\text{TOP}_{k}(\boldsymbol{S})$                                                                                             & $\begin{aligned} \boldsymbol{X}^{\prime}=\boldsymbol{X}_{\mathrm{idx}} \odot \boldsymbol{S}_{\mathrm{idx}}; \quad \boldsymbol{A}^{\prime}=\boldsymbol{A}_{\mathrm{idx}, \mathrm{idx}}\end{aligned}$                                                                                                                                                                                                                                                                       \\ 
\begin{tabular}[c]{@{}l@{}}AttPool~\cite{attpool}\tnote{1} \end{tabular}      & $\boldsymbol{S}=  softmax\left(\boldsymbol{X}\boldsymbol{W}\right)$                                                                                            & $\mathrm{idx}=\text{TOP}_{k}(\boldsymbol{S})$   & $\begin{aligned} \boldsymbol{X}^{\prime}= \boldsymbol{A}_{\mathrm{idx}}( \boldsymbol{X} \odot \boldsymbol{S}); \quad \boldsymbol{A}^{\prime}=\boldsymbol{A}_{\mathrm{idx}} \boldsymbol{A} \boldsymbol{A}_{\mathrm{idx}}^{T}   \end{aligned}$                                                                                                                                                                                                                                                                       \\ 
\begin{tabular}[c]{@{}l@{}}ASAP~\cite{asap}\end{tabular}            & $\boldsymbol{S} = LEConv(\boldsymbol{X^{c}},\boldsymbol{A})$                                                                                                           &  $\mathrm{idx}=\text{TOP}_{k}(\boldsymbol{S})$                                                                                              & $\begin{aligned} \boldsymbol{X}^{\prime} =\boldsymbol{X}^{c}_{\mathrm{idx}} \odot \boldsymbol{S}_{\mathrm{idx}}; \quad \boldsymbol{A}^{\prime}=\boldsymbol{A}_{\mathrm{idx}} \boldsymbol{A} \boldsymbol{A}_{\mathrm{idx}}^{T}   \end{aligned}$                                                                                                                                                                                                                                                               \\ 
\begin{tabular}[c]{@{}l@{}}HGP-SL~\cite{hgp-sl}\end{tabular}        & $\boldsymbol{S}=\left\|\left(\boldsymbol{I}-\boldsymbol{D}^{-1} \boldsymbol{A}\right) \mathbf{X}\right\|_{1}$                           & $\mathrm{idx}=\text{TOP}_{k}(\boldsymbol{S})$                                                                                              & $\left\{ \begin{aligned}&\boldsymbol{X}^{\prime}=\boldsymbol{X}_{\mathrm{idx}} \odot \boldsymbol{S}_{\mathrm{idx}}; \quad \boldsymbol{A}^{\prime}=\boldsymbol{A}_{\mathrm{idx}, \mathrm{idx}} \\& \boldsymbol{A}^{\prime \prime}_{ij} = sparsemax(\sigma(\overrightarrow{\mathbf{a}}[\boldsymbol{X}^{\prime}(i,:) \|\boldsymbol{X}^{\prime}(j,:)]^{\top})+\lambda \cdot \boldsymbol{A}^{\prime}_{ij} \end{aligned}\right.$                                                                                  \\ 

\begin{tabular}[c]{@{}l@{}}VIPool~\cite{vip-pool}\end{tabular}      & $\left\{\begin{aligned}&\boldsymbol{P} = \frac{1}{t} \sum_{h=1}^{t} \left(\tilde{\boldsymbol{D}}^{-\frac{1}{2}} \tilde{\boldsymbol{A}} \tilde{\boldsymbol{D}}^{-\frac{1}{2}} \right)^{h} \boldsymbol{W}^{h} \text{MLP}(\boldsymbol{X}) \\& \boldsymbol{S} = \sigma(\text{MLP}(\text{MLP}(\boldsymbol{X}, \boldsymbol{P})) ) \end{aligned}\right.$ & $\mathrm{idx}=\text{TOP}_{k}(\boldsymbol{S})$                                                                                              & $\left\{ \begin{aligned} & \boldsymbol{X}^{\prime}=\boldsymbol{X}_{\mathrm{idx}} \odot \boldsymbol{S}_{\mathrm{idx}} \\& \boldsymbol{A}^{\prime}= softmax(\boldsymbol{A}_{\mathrm{idx}}) \boldsymbol{A} softmax( \boldsymbol{A}_{\mathrm{idx}})^{T} \end{aligned} \right.$                                                                                                                                                                                                                                                                                   \\ 
\begin{tabular}[c]{@{}l@{}}RepPool~\cite{rep-pool}\end{tabular}     & $\boldsymbol{S}=\sigma\left(\boldsymbol{D}^{-1} \boldsymbol{A}  \boldsymbol{X} \mathbf{p} /\left\|\mathbf{p}\right\|_{2}\right)$                                                    & $\mathrm{idx}=\text{SEL}_{k}(\boldsymbol{S})$                                                                                              & $\left\{ \begin{aligned}&\boldsymbol{B}=\boldsymbol{X} \boldsymbol{W}_{b}(\boldsymbol{X}_{\mathrm{idx}})^{T}  \\ &\boldsymbol{X}^{\prime}=(softmax(\boldsymbol{B} \odot \boldsymbol{M}))^{T} (\boldsymbol{X} \odot(\boldsymbol{S} ))\\ &\boldsymbol{A}^{\prime}=(softmax(\boldsymbol{B} \odot \boldsymbol{M}))^{T} \boldsymbol{A}(softmax(\boldsymbol{B} \odot \boldsymbol{M})) \end{aligned} \right.$ \\ 

\begin{tabular}[c]{@{}l@{}}UGPool~\cite{uniform-pooling}\end{tabular}     & $\boldsymbol{S}=\sigma(\boldsymbol{X} \mathbf{p} /\left\|\mathbf{p}\right\|_{2}) $                                                    & $\mathrm{idx}=\text{1DPool}(\text{rank}(\boldsymbol{S}))$                                                                                              &  $\begin{aligned} \boldsymbol{X}^{\prime}=\boldsymbol{X}_{\mathrm{idx}} \odot \boldsymbol{S}_{\mathrm{idx}}; \quad \boldsymbol{A}^{\prime}=\boldsymbol{A}_{\mathrm{idx}, \mathrm{idx}} + \boldsymbol{A}_{\mathrm{idx}, \mathrm{idx}}^{2} \end{aligned}$  \\ 

\begin{tabular}[c]{@{}l@{}}GSAPool~\cite{gsapool}\end{tabular}      & $\left\{\begin{aligned}&\boldsymbol{S}_{1}=\sigma\left(\tilde{\boldsymbol{D}}^{-\frac{1}{2}} \tilde{\boldsymbol{A}} \tilde{\boldsymbol{D}}^{-\frac{1}{2}} \boldsymbol{X}\boldsymbol{W}\right)  \\& \boldsymbol{S}_{2} = \sigma(\text{MLP}(\boldsymbol{X})) \\& \boldsymbol{S} = \alpha \boldsymbol{S}_{1} + (1- \alpha)\boldsymbol{S}_{2}\end{aligned}\right.$ & $\mathrm{idx}=\text{TOP}_{k}(\boldsymbol{S})$                                                                                              & $ \boldsymbol{X}^{\prime}=(\boldsymbol{A}\boldsymbol{X}\boldsymbol{W})_{\mathrm{idx}} \odot \boldsymbol{S}_{\mathrm{idx}}; \quad \boldsymbol{A}^{\prime}=\boldsymbol{A}_{\mathrm{idx}, \mathrm{idx}}$                                                                                                                                                                                                                                                                                    \\ 

\begin{tabular}[c]{@{}l@{}}CGIPool~\cite{cgi-pool}\end{tabular}      & $\left\{\begin{aligned}&\boldsymbol{S}_{r}=\sigma\left(\tilde{\boldsymbol{D}}^{-\frac{1}{2}} \tilde{\boldsymbol{A}} \tilde{\boldsymbol{D}}^{-\frac{1}{2}} \boldsymbol{X}\boldsymbol{W}_{r}\right)  \\& \boldsymbol{S}_{f}=\sigma\left(\tilde{\boldsymbol{D}}^{-\frac{1}{2}} \tilde{\boldsymbol{A}} \tilde{\boldsymbol{D}}^{-\frac{1}{2}} \boldsymbol{X}\boldsymbol{W}_{f}\right) \\& \boldsymbol{S} = \sigma( \boldsymbol{S}_{r} - \boldsymbol{S}_{f}) \end{aligned}\right.$ & $\mathrm{idx}=\text{TOP}_{k}(\boldsymbol{S})$                                                                                              & $\begin{aligned} \boldsymbol{X}^{\prime}=\boldsymbol{X}_{\mathrm{idx}} \odot \boldsymbol{S}_{\mathrm{idx}}; \quad \boldsymbol{A}^{\prime}=\boldsymbol{A}_{\mathrm{idx}, \mathrm{idx}}\end{aligned} $                                                                                                                                                                                                                                                                                   \\ 

\begin{tabular}[c]{@{}l@{}}TAPool~\cite{TAPool}\end{tabular}      & $\left\{\begin{aligned}&\boldsymbol{S}_{l}=softmax\left( \frac{1}{n} ((\boldsymbol{X} \boldsymbol{X}^{T}) \odot (\tilde{\boldsymbol{D}}^{-1} \tilde{\boldsymbol{A}})) \boldsymbol{1}_{n} \right)  \\& \boldsymbol{S}_{g}= softmax \left(\tilde{\boldsymbol{D}}^{-1} \tilde{\boldsymbol{A}} \boldsymbol{X} \mathbf{p} \right)  \\& \boldsymbol{S} =  \boldsymbol{S}_{l} + \boldsymbol{S}_{g} \end{aligned}\right.$ & $\mathrm{idx}=\text{TOP}_{k}(\boldsymbol{S})$                                                                                              & $\begin{aligned} \boldsymbol{X}^{\prime}=\boldsymbol{X}_{\mathrm{idx}} \odot \boldsymbol{S}_{\mathrm{idx}}; \quad \boldsymbol{A}^{\prime}=\boldsymbol{A}_{\mathrm{idx}, \mathrm{idx}}\end{aligned} $                                                                                                                                                                                                                                                                                   \\ 

\begin{tabular}[c]{@{}l@{}}IPool~\cite{ipool}\tnote{2}\end{tabular}          & $\boldsymbol{S}=\left\|\left(\boldsymbol{I}-\frac{1}{t} \sum_{h=1}^{t} (\bar{\boldsymbol{D}}^{h})^{-1} \bar{\boldsymbol{A}}^{h} \right) \boldsymbol{X}\right\|_{2} $                                                   & $\mathrm{idx}=\text{TOP}_{k}(\boldsymbol{S})$  & $ \left\{ \begin{aligned}&\boldsymbol{X}^{\prime}=\boldsymbol{X}_{\mathrm{idx}}, \\&\boldsymbol{A}^{\prime}_{i j}=\lambda\left(\boldsymbol{A}+\boldsymbol{I}\right)_{\mathrm{idx}[i], \mathrm{idx}[j]} + (1- \lambda) \boldsymbol{O}_{ij}  \end{aligned}\right. $                                                                                                                                                                                                                          \\ \bottomrule
\end{tabular}%
}\begin{tablenotes}
  \scriptsize  
  \item[1] This manuscript only presents the global attention mechanism to generate scores, and there is  also a local attention mechanism introduced in ~\cite{attpool}.  
  \item[2] This manuscript only presents the greedy  IPool strategy in this table, and there is  also a local IPool strategy  introduced in ~\cite{ipool}. 
  \end{tablenotes}
  \end{threeparttable}
  \begin{minipage}{\linewidth}\scriptsize
\textbf{Notations:} $\boldsymbol{X}^{\prime} \in \mathbb{R}^{ k \times c}$ and  $\boldsymbol{A}^{\prime} \in \{0, 1\}^{ k \times k}$  are the adjacency matrix and feature matrix for the new graph, respectively;  $\boldsymbol{X^{c}}$ is the cluster representation matrix calculated by a new variant of self-attention; $\tilde{\boldsymbol{A}} \in \{0, 1\}^{ n \times n}$ is the adjacency matrix with self-connections (i.e. $\tilde{\boldsymbol{A}} = \boldsymbol{A}+\boldsymbol{I}$); $\tilde{\boldsymbol{D}} \in \mathbb{R}^{ n \times n} $ is the degree matrix of $\tilde{\boldsymbol{A}}$; $\boldsymbol{D} \in \mathbb{R}^{ n \times n} $ is the degree matrix of $\boldsymbol{A}$; $\bar{\boldsymbol{A}}^{h} \in \mathbb{R}^{ n \times n}$ is the matrix where diagonal values corresponding to the h-hop circles have been removed; $\bar{\boldsymbol{D}}^{h} \in \mathbb{R}^{ n \times n}$ is the corresponding degree matrix of $\bar{\boldsymbol{A}}^{h}$; $\boldsymbol{W} \in \mathbb{R}^{ c \times 1}$; $\boldsymbol{W}_{r} \in \mathbb{R}^{ c \times 1}$; $\boldsymbol{W}_{f} \in \mathbb{R}^{ c \times 1}$ and  $\boldsymbol{W}_{b} \in \mathbb{R}^{ c \times c}$ are the learnable weight matrices;  $\mathbf{p} \in \mathbb{R}^{c}$ and $\mathbf{a} \in \mathbb{R}^{ 1 \times 2c}$ are the trainable projection vectors; $\boldsymbol{I}$ is the identity matrix;  $\lambda $ is a trade-off parameter;  $\alpha$ is a user-defined hyperparameter; $\mathbf{1} \in \mathbb{R}^{n} $ is a vector with all elements being 1; $\boldsymbol{M}  \in \mathbb{R}^{ n \times k} $ is a masking matrix; $\boldsymbol{O}$ is the matrix to measure the overlap of neighbors between nodes; $\sigma$ is the activation function (e.g. $tanh$);  $\odot$ is the broadcasted elementwise product; $\text{MLP}$ is a multi-layer perceptron; $\text{1DPool}(\cdot)$ is the normal pooling on one-dimensional data; $LEConv$ is Local Extrema Convolution which helps capture local extremum information; $\text{SEL}_{k}$ is the algorithm to select nodes one by one.
\end{minipage}
\end{table*}

We first summarize previous node drop pooling methods to give an in-depth analysis. Specifically, we propose a universal and modularized framework to describe the process of node drop pooling, see Fig.~\ref{fig:framework}. We deconstruct node drop graph pooling with three disjoint modules: \textbf{1) Score generator.} Given an input graph, the score generator calculates the significance scores for each node. \textbf{2) Node selector.} Node selector selects the nodes with the top-k significance scores.  \textbf{3) Graph coarser.} With the selected nodes, a  new graph coarsened from the original one is obtained by learning the feature matrix and the adjacency matrix. The process can be formulated as follows:
\begin{equation}
\begin{aligned}
&\underbrace{\boldsymbol{S}^{(l)}=\text{SCORE}(\boldsymbol{X}^{(l)}, \boldsymbol{A}^{(l)})}_{\text {\normalsize Score generator}} ;   \quad \underbrace{\mathrm{idx}^{(l+1)}=\text{TOP}_{k}(\boldsymbol{S}^{(l)})}_{\text{\normalsize Node selector}} ;  \\
&\underbrace{\boldsymbol{X}^{(l+1)}, \boldsymbol{A}^{(l+1)}=\text{COARSOR}(\boldsymbol{X}^{(l)}, \boldsymbol{A}^{(l)},\boldsymbol{S}^{(l)},\mathrm{idx}^{(l+1)})}_{\text{\normalsize Graph coarser}}, 
\end{aligned} 
\end{equation}
where functions $\text{SCORE}$, $\text{TOP}_{k}$, and $\text{COARSOR}$ are specially designed by each method for the score generator, node selector, and graph coarsor, respectively. $\boldsymbol{S}^{(l)}\in \mathbb{R}^{ n \times 1}$ indicates the significance scores; $\text{TOP}_{k}$ ranks values and returns the indices of the largest $\lceil {k}\times {n} \rceil$ values in $\boldsymbol{S}^{(l)}$; $\mathrm{idx}^{(l+1)}$ indicates the reserved node indexes for the new graph, and $l$ as well as $l+1$ indicates the layer numbers. 

Accordingly, we present how the twelve typical node drop pooling models fit into our proposed framework in Table~\ref{tab:schemes}. It is observed that the vast majority of node drop pooling methods focus on advancing the score generator and graph coarsor modules, but ignore the design of the node selector module. However, a simple node selector highlights nodes with similar features or structures during training without considering the node-feature and the graph-structure diversities in graphs. 

\subsection{Proposed MID}

In contrast to the previous node drop pooling models, we mainly focus on the node selector module. Based on this, we propose MID, which is a plug-and-play scheme for improving node drop pooling. As shown in Fig.~\ref{fig:model}, MID consists of   \textbf{1) a multidimensional score space},  \textbf{2) a flipscore operation}, and  \textbf{3) a dropscore operation}. With the above three components appropriately integrated, MID is enabled to mitigate the issues of ignoring node-feature and graph-structure diversities in modern pooling models. Each component is discussed and justified  analytically in detail in the following subsections.

\subsubsection{The Multidimensional Score Space}
\label{sec:multi-score}

We first extend the original one-dimensional score space $\boldsymbol{S}^{(l)}\in \mathbb{R}^{ n \times 1}$  to a multidimensional score space, with one dimension for one view of the nodes, thus comprehensively depicting node information:
\begin{equation}
    \boldsymbol{S}_\text{multi}^{(l)}=\text{SCORE}(\boldsymbol{X}^{(l)}, \boldsymbol{A}^{(l)}),
\label{eq:multi-score}
\end{equation}
where $\boldsymbol{S}_\text{multi}^{(l)}\in \displaystyle \mathbb{R}^{ n \times h}$ is the significance score matrix. Different from $\boldsymbol{S}^{(l)}\in \mathbb{R}^{ n \times 1}$ in the base pooling models, $\boldsymbol{S}_\text{multi}^{(l)}$ suggests that for each node, its signiﬁcance is to be evaluated by different scalars. Take TopKPool~\cite{graph-u-net} as an example, we broaden the trainable projection vector $\mathbf{p} \in \mathbb{R}^{c}$ to a projection matrix  $\boldsymbol{P} \in \mathbb{R}^{c \times h}$, and then the score generator would predict $h$ different scalars (a h-dimensional vector) for each node.

\textbf{Intuition.} Previous methods mainly use a scalar quantity to represent the importance of each node. However, due to the complexity of a graph, one scalar is not sufficient to depict the significance of a node from different views. Therefore, we extend the importance score for each node from a scalar to a multidimensional vector to evaluate the significance of nodes from multi-views.

\begin{figure}[!t] 
\begin{center}
\setlength{\abovecaptionskip}{-0.1cm}
\includegraphics[width=1.0\linewidth]{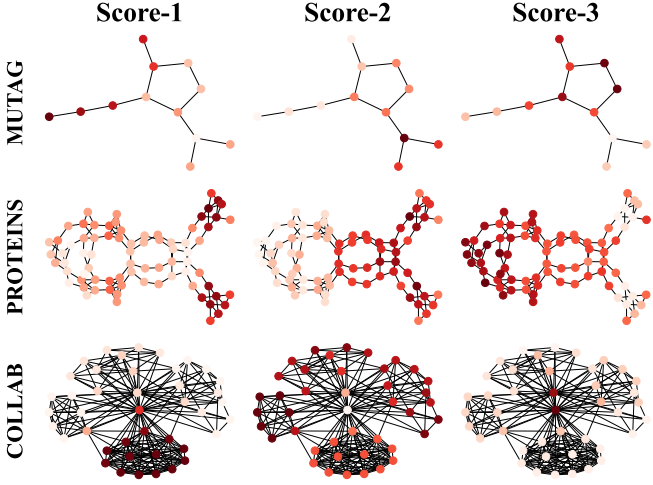}
\end{center}
\caption{Illustration of the \textbf{multidimensional} score mechanism. Node color indicates its score. Darker the shade, higher the score. Score-$n$ refers to the $n-$th value of the score vector. }
\label{fig:multi-diversity}
\end{figure}

\textbf{Case Study.} To investigate \textit{whether the multidimensional score operation can assist models in evaluating the significance of nodes from multi-views}, we conduct experiments on three benchmark datasets (MUTAG, PROTEINS, and COLLAB), involving different graph sizes and domains (social and bio-chemical),  by using the SAGPool model~\cite{sagpool}. Fig.~\ref{fig:multi-diversity} illustrates the scores predicted by the multidimensional score operation with $h =3$.  We can observe that scores from different views highlight different nodes in the graph, which indicates our multidimensional score operation encourages models to capture the diversity in the graph.

\begin{figure}[!t] 
\setlength{\abovecaptionskip}{-0.1cm}   
\begin{center}
\includegraphics[width=1.0\linewidth]{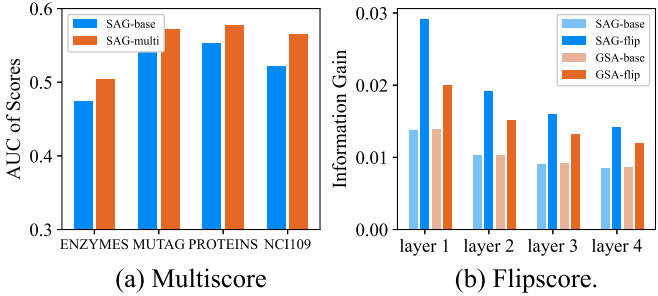}
\end{center}
\caption{(a) Score correctness. SAG-base and SAG-multi refer to SAGPool model with or without a multidimensional score space, respectively. (b) Information gain. SAG-flip and GSA-flip refer to SAGPool model with flipscore operation and GSAPool model with flipscore operation, respectively.}
\label{fig:case_study_multi}
\end{figure}

\textbf{Analytical Verification.} To validate the effectiveness of the multidimensional score space,  We further explore \textit{whether the multidimensional score evaluates the importance of the nodes more accurately} by calculating the score correctness~\cite{understand-attention, attention-accuracy} on four benchmark datasets (ENZYMES, MUTAG, PROTEINS, and NCI109) with SAGPool~\cite{sagpool} model. Specifically, after training a model, we remove node $i$ from a graph and compute an absolute difference from prediction $y$ for the original graph:
\begin{equation}
s_{i}^{GT}=\frac{\left|y_{i}-y\right|}{\sum_{j=1}^{n}\left|y_{j}-y\right|},
\end{equation}
where $s_{i}^{GT}$ is the ground truth score and $y_{i}$ is a model’s prediction for the graph without node $i$. After obtaining the ground truth scores and calculated scores in the first pooling layer on the test dataset, we evaluate score correctness using area under the ROC curve (AUC) following~\cite{understand-attention}, which allows us to evaluate the ranking of scores rather than their absolute values. As shown in Fig.~\ref{fig:case_study_multi} (a), models with a multidimensional score space consistently generate more accurate scores for nodes, which confirms that a multidimensional vector used to depict the significance of nodes benefits the evaluation.

\begin{figure}[!t] 
\setlength{\abovecaptionskip}{-0.1cm}   
\begin{center}
\includegraphics[width=1.0\linewidth]{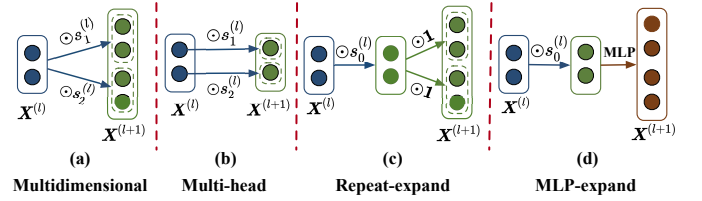}
\end{center}
\caption{Experiments to validate the effect of the dimension of embedding vectors. We consider a node with 2 features. The input feature matrix of layer $l$ is  $\boldsymbol{X}^{(l)} \in \mathbb{R}^{ 1 \times 2}$; the output feature matrix in (a), (c) and (d) is  $\boldsymbol{X}^{(l+1)} \in \mathbb{R}^{ 1 \times 4}$; and the output feature matrix in (b) is $\boldsymbol{X}^{(l+1)} \in \mathbb{R}^{ 1 \times 2}$.}
\label{fig:multi-score}
\end{figure}

\begin{figure}[!t] 
\begin{center}
\setlength{\abovecaptionskip}{-0.1cm}
\includegraphics[width=1.0\linewidth]{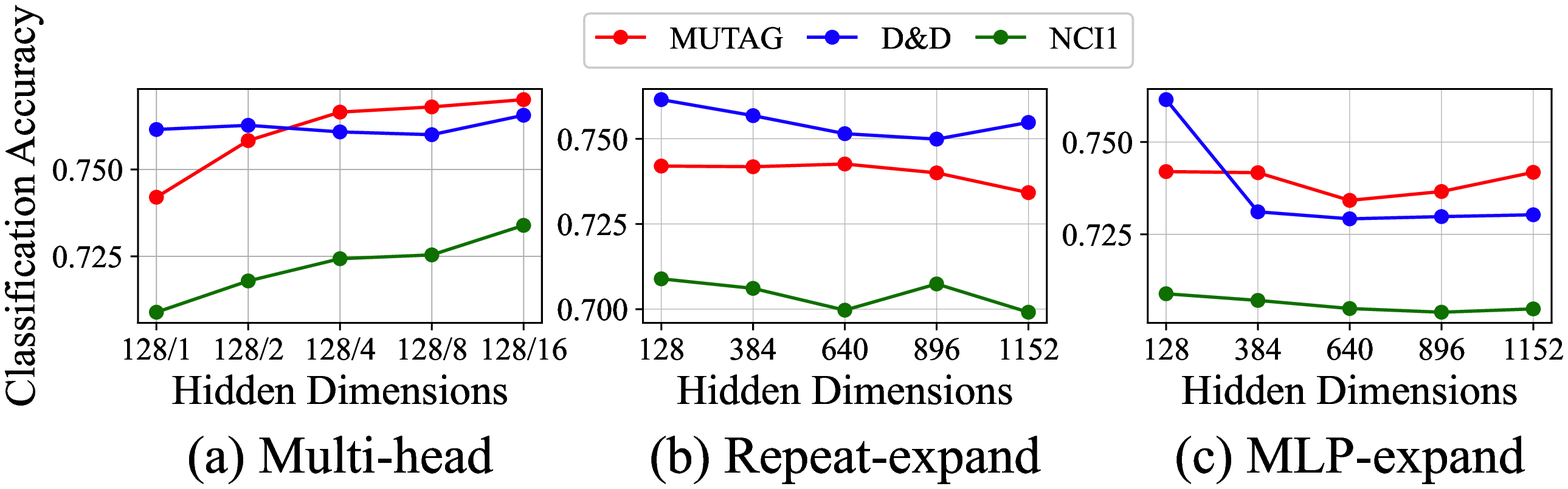}
\end{center}
\caption{Accuracy results varying with different hidden dimensions. 128/2 means that the dimension of the embedding vector $c = 128$, and the dimension of score $h =2$. }
\label{fig:hidden}
\end{figure}

\textbf{Additional Dimension Analysis.}  We utilize scores generated by Eq.~(\ref{eq:multi-score}) to generate the pooled feature map.   Let $\boldsymbol{S}_\text{multi}^{(l)}=\{\mathbf{s}_1^{(l)}\|\mathbf{s}_2^{(l)}\|,...,\|\mathbf{s}_h^{(l)} \}$, and then the new feature map of nodes is defined as: 
\begin{equation}
\{(\boldsymbol{X}^{(l)}\odot \mathbf{s}_1^{(l)})\|(\boldsymbol{X}^{(l)}\odot \mathbf{s}_2^{(l)})\|, ...,\|(\boldsymbol{X}^{(l)}\odot \mathbf{s}_h^{(l)})\},
\end{equation}
where $\mathbf{s}_h^{(l)} \in \mathbb{R}^{n \times 1}$ is one column of  $\boldsymbol{S}_\text{multi}^{(l)}$,  and $\|$ indicates the vector concatenation. As shown in Fig.~\ref{fig:multi-score} (a), the above multidimensional score space operation would enlarge the dimension of embedding vectors~\footnote{The dimension of embedding vectors would be downscaled to the standard size after the propagation of the vectors in the next GCN layer.}. 
Therefore, this section gives an in-depth study on \textit{ how the embedding dimension affects the performance in the graph classification task.}   
For a fair comparison, we design three experiments, see Fig.~\ref{fig:multi-score}.  \textbf{1) Multi-head} (in Fig.~\ref{fig:multi-score} (b)).  We adopt the multi-head mechanism in GAT~\cite{gat} to achieve the goal to decrease the dimension after calculating the score. Let $\boldsymbol{X}^{(l)}=\{\boldsymbol{x_1}^{(l)}\|\boldsymbol{x_2}^{(l)}\|...,\|\boldsymbol{x_h}^{(l)} \}$, and then the new feature map of nodes is defined as:  
\begin{equation}
\{(\boldsymbol{x_1}^{(l)}\odot \mathbf{s}_1^{(l)})\|(\boldsymbol{x_2}^{(l)}\odot \mathbf{s}_2^{(l)})\|,...,\|(\boldsymbol{x_h}^{(l)}\odot \mathbf{s}_h^{(l)})\},
\end{equation} 
where $\boldsymbol{x}^{(l)} \in \mathbb{R}^{n \times d}$, and $d = \lceil {c}/ {h} \rceil$.  Besides, decreasing the dimension, we also test what would happen if we increase the dimension by other methods. \textbf{2) Repeat-expand} (in Fig.~\ref{fig:multi-score} (c)). We stack the embedding vectors to raise up the dimension in the pooling layer, and then the new features of the nodes are defined as:  \begin{equation}
\{(\boldsymbol{X}^{(l)}\odot \boldsymbol{S}^{(l)})\|(\boldsymbol{X}^{(l)}\odot \boldsymbol{S}^{(l)})\|,...,\|(\boldsymbol{X}^{(l)}\odot \boldsymbol{S}^{(l)})\}.\end{equation} \textbf{3) MLP-expand} (Fig.~\ref{fig:multi-score} (d)). 
We also try to raise the embedding dimension in base models by adding one layer of the multi-layer perceptron after the gate operation ($\boldsymbol{X}^{(l)}\odot \boldsymbol{S}^{(l)}$). We conduct the above three experiments on three benchmark datasets (MUTAG, DD, and NCI1) with the SAGPool model as a baseline model, and the samples of these datasets are 188, 1,178, and 4,110, respectively. The detailed results in Fig.~\ref{fig:hidden} (b) and (c) illustrate that raising the hidden dimensions in MUTAG, DD, and NCI1 datasets contributes nothing to the accuracy improvement, and what's worse, the accuracy decades as the dimension increases. And in Fig.~\ref{fig:hidden} (a), it is demonstrated that our multidimensional score space operation contributes to the accuracy improvement even if the hidden dimensions are not raised.

\subsubsection{The Flipscore Operation}
 
After getting a multidimensional score $\boldsymbol{S}_\text{multi}^{(l)}$, we entail the scores generated by models to extract as many diverse nodes as possible. Specifically, the flipscore operation yields the absolute value of each element in score $\boldsymbol{S}_\text{multi}^{(l)}$:
\begin{equation}
\boldsymbol{S}_\text{flip}^{(l)}(i,j) = \left| \boldsymbol{S}_\text{mutli}^{l}(i,j) \right|,
\label{eq:flip-score}
\end{equation}
where $ \boldsymbol{S}_\text{flip}^{(l)} \in \displaystyle \mathbb{R}^{ n \times h} $ is the resulting score matrix, and $\left| \cdot \right|$ is the absolute value of the argument. Note that the flipscore operation will not influence the node feature updating of graph coarsor. We adopt this operation in training, validation, and test phases.
 
\textbf{Intuition.}  Previous methods tend to highlight similar and significant nodes instead of representative and significant nodes. From an information theory standpoint, similar features do not add extra information to the feature hierarchy, and therefore should be possibly suppressed~\cite{ipool}. According to the summary in Table~\ref{tab:schemes}, the generated scores of most previous models range from -1 to 1 after using Tanh~\cite{sagpool,graph-u-net,gsapool}. Therefore, our flipscore operation could highlight nodes with extremely different scores in the original space, encouraging node drop pooling to capture nodes with dissimilar features. Intuitively, dissimilar features lead to more information gains which experimentally contribute to a better graph classification~\cite{hgp-sl}.

\begin{figure}[!t] 
\begin{center}
\setlength{\abovecaptionskip}{-0.1cm}
\includegraphics[width=1.0\linewidth]{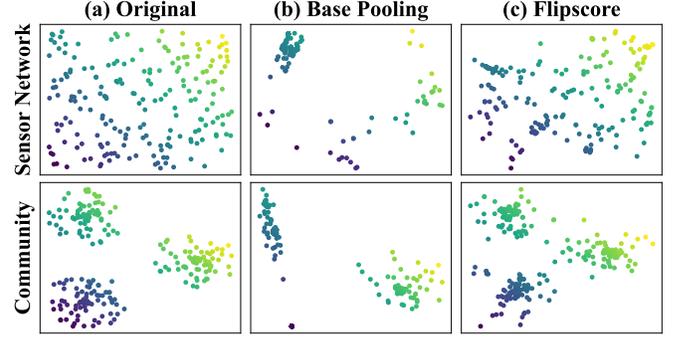}
\end{center}
\caption{Illustration of the \textbf{flipscore} mechanism. Reconstruction results of two synthetic graphs, compared to base node drop pooling methods.}
\label{fig:case-flip}
\end{figure}

\textbf{Case Study.} To investigate \textit{whether the flipscore operation can assist models to capture the node-feature diversity}, we conduct experiments on two synthetic graphs (Community graph and David sensor network) from the PyGSP library~\cite{pygsp}, by using the SAGPool model~\cite{sagpool}.  The experimental settings follow the proposals in the previous works~\cite{mincut, gmt}.  Fig.~\ref{fig:case-flip} reports the original graph feature map (the node features are the 2-D coordinates of the nodes) and the reconstruction by the base pooling method (SAGPool) and pooling method with flipscore. We can observe that SAGPool fails to recover the original graph signal, while SAGPool with flipscore yields better results in both cases, which indicates our flipscore encourages models to capture the node-feature diversity in the graph.

\textbf{Analytical Verification.} We further give a verification in terms of information gain. Specifically, we compute information gained from surrounding after one aggregation and pooling operation by Kullback–Leibler divergence~\cite{kl-div}, following the previous study~\cite{cs-gnn}:
\begin{equation}
D_{K L}\left(U \| C \right)=\int_{\mathcal{X}} U(\boldsymbol{x}) \cdot \log \frac{U(\boldsymbol{x})}{C(\boldsymbol{x})} d \boldsymbol{x},
\end{equation}
where $\mathcal{X}$ is the normalized feature space, $C$ is the probability density function (PDF) estimated by the term $\breve{\boldsymbol{c}}_{v}$, which is the ground-truth vector of node $v$, $U$ is the probability density function (PDF) estimated by the term $\sum_{v_{j} \in \mathcal{N}_{v_i}} a_{i j} \cdot \breve{c}_{v}$, $\mathcal{N}_{v_{i}}$ is the set of neighbors of node $v_{i}$, and $ a_{i j}$ is the coefficient of node $v_{j}$ to node $v_{i}$. According to the proof (the detailed proof can be found in ~\cite{lookhops, cs-gnn}), $D_{K L}$ is positively correlated to :
\begin{equation}
\frac{\left\|\sum_{v \in \mathcal{V}}\left(\sum_{v^{\prime} \in \mathcal{N}_{v}}\left(x_{v}-x_{v^{\prime}}\right)\right)^{2}\right\|_{1}}{|\mathcal{E}| \cdot c}.
\end{equation}
Therefore, we calculate the pooling information gain of selected nodes by :

\begin{equation}
\lambda_{\text{SEL}} =\frac{\left\|\sum_{v \in \mathcal{V}_{\text{SEL}}}\left(\sum_{v^{\prime} \in \mathcal{N}_{v}}\left(x_{v}-x_{v^{\prime}}\right)\right)^{2}\right\|_{1}}{|\mathcal{E}| \cdot c}, 
\label{eq:kl}
\end{equation}
where $\mathcal{V}_{\text{SEL}}$ is the set of selected nodes, and $\lambda_{
\text{SEL}}$ is the information gain  during one-time pooling operation. We conduct our experiments on the MUTAG dataset by applying SAGPool and GSAPool models with or without the flipscore operation. As shown in Fig.~\ref{fig:case_study_multi} (b), our flipscore operation significantly improves the information gain in all cases, which confirms that the flipscore operation promotes models to maintain more diverse node features.

\subsubsection{The Dropscore Operation}

The dropscore operation is proposed to force node drop pooling models to notice as many substructures of graphs as possible. Specifically, dropscore operation randomly drops out several nodes with a certain rate in the graph during training only when selecting the top-k scores:
\begin{equation}
 \boldsymbol{S}_\text{drop}^{(l)} = \boldsymbol{I}_{\lceil {p_s}\times {n}\rceil} \boldsymbol{S}_\text{multi}^{(l)},
 \label{eq:drop-score}
\end{equation}
where $\boldsymbol{S}_\text{drop}^{l} \in \displaystyle \mathbb{R}^{ n \times h}$ is the resulting score matrix, $ \boldsymbol{I}_{\lceil {p_s}\times {n}\rceil} \in \displaystyle \mathbb{R}^{ n \times n}$ is a matrix generated by randomly dropping $\lceil {p_s}\times {n} \rceil$ none-zero elements of a unit matrix with $n$ dimensions, $\lceil \cdot \rceil$ is the operation of rounding up,  and  $p_s$ is the score dropping rate. We adopt this operation in the training phase.

\textbf{Intuition.} Note that through GNNs, nodes that are directly connected tend to share similar information~\cite{birds,sgcn}. Therefore, models generate similar scores for nearby nodes, which is also observed in the previous study~\cite{understand-attention}. Under this condition, models may be stuck into significant local structures and select redundant nodes, thus ignoring significant nodes from other substructures and losing structure information. Therefore, we devise dropscore operation, which randomly drops out a certain rate of nodes during training, expecting that models do not focus on one local substructure by removing several nodes in the local substructure.

\begin{figure}[!t] 
\setlength{\abovecaptionskip}{-0.1cm}   
\begin{center}
\includegraphics[width=1.0\linewidth]{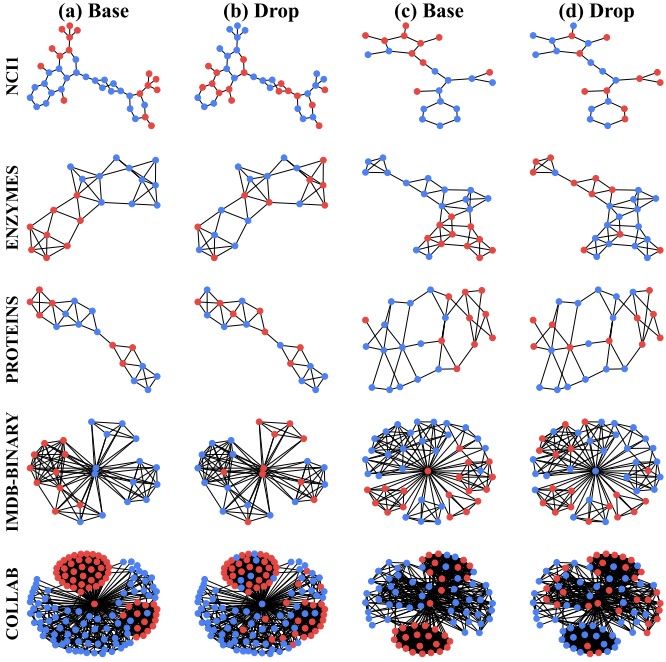}
\end{center}
\caption{Illustration of the \textbf{dropscore} mechanism. Visualization of node selection results with (\textbf{Drop}) and without (\textbf{Base}) the dropscore operation. Reserved nodes are highlighted in {\color[HTML]{FE0000} red}.}
\label{fig:case_study_drop}
\end{figure}

\textbf{Case Study.} To investigate \textit{whether the dropscore operation can assist models to cover more significant substructures}, we conduct experiments on five benchmark datasets (NCI1, ENZYMES, PROTEINS, IMDB-BINARY, and COLLAB), involving different graph sizes and domains (social and bio-chemical),  with the SAGPool model~\cite{sagpool}.  40\% nodes are selected in the first pooling layer and are highlighted in red.   As shown in Fig.~\ref{fig:case_study_drop}, SAGPool (\textbf{Base}) is likely to select nodes concentrated in the same area, which supports our above intuition. Therefore, important information of other parts might be neglected. With our dropscore operation (\textbf{Drop}), selected nodes are distributed in different substructures covering the whole graph, confirming that our method encourages models to maintain more diverse structure information.

\subsection{Discussion of MID}

In this subsection, we first present an in-depth analysis regarding the propositions of the trapped scores and the graph permutation equivariance, and then demonstrate that our proposed MID could enable the backbone pooling models to be more powerful in expressiveness.

\subsubsection{Trapped Scores}

\begin{remark}

The scores generated by based pooling models tend to be stuck in the local structure, and they are referred to as \textbf{trapped scores}.
	
\end{remark}

As shown in Fig.~\ref{fig:case_study_drop}, the selected nodes (highlighted in red) predicted by pooling models tend to be stuck in the local structure. To illustrate the above phenomenon, we take the TopKPool~\cite{graph-u-net,topkpool} model as an example. TopKPool predicts the scores through:
\begin{equation}
    \mathbf{S}=\boldsymbol{X} \frac{\mathbf{p}}{\|\mathbf{p}\|_{2}},
\end{equation}
where $ \mathbf{p} \in \displaystyle \mathbb{R}^{ n \times 1}$ is a  learnable vector, and $ \mathbf{S} \in \displaystyle \mathbb{R}^{ n \times 1}$ is the predicted scores for nodes. Given two feature vectors $\mathbf{x}_1$ and $\mathbf{x}_2$ for nodes $u$ and $v$ respectively, the scores of two nodes are:
\begin{equation}
s_{1}= \mathbf{x}_{1} \frac{\mathbf{p}^T}{\|\mathbf{p}\|_{2}},	\quad s_{2}= \mathbf{x}_{2} \frac{\mathbf{p}^T}{\|\mathbf{p}\|_{2}}.
\end{equation}
Then, to evaluate the distance between the two scores, we have:
\begin{align}
\label{eq:trap}
    \left|s_{1}-s_{2}\right|  = \left|  \mathbf{x}_{1} \frac{\mathbf{p}^T}{\|\mathbf{p}\|_{2}}  - \mathbf{x}_{2} \frac{\mathbf{p}^T}{\|\mathbf{p}\|_{2}}  \right| \le \| \mathbf{x}_{1} - \mathbf{x}_{2}\|_{2}.
\end{align}
Assume that linked nodes in the same local structure have similar features. Such an assumption is common in most real-world networks~\cite{birds}. Therefore, if  nodes $u$ and $v$ are in the same local structure, that is, $\left\|\mathbf{x}_{1}-\mathbf{x}_{2}\right\| \rightarrow 0$, then we get:
\begin{equation}
	\left|s_{1}-s_{2}\right| \rightarrow 0.
\end{equation}

The above analysis demonstrates that the selected nodes (correspond to trapped scores) of TopKPool models are stuck in the local structure. Moreover, this finding  could be applied to other cases, \textit{e.g.}, SAGPool~\cite{sagpool}, GASPool~\cite{gsapool}, and ASAP~\cite{asap}.

\subsubsection{Graph Permutation Equivariance}

\begin{remark}
	Suppose that the backbone graph pooling model is graph permutation equivariant, and then the model combined with our MID  is still graph permutation equivariant.
\end{remark}

Graph pooling generates isomorphic pooled graphs after graph permutation, which is defined as graph permutation equivariance~\cite{asap}. As formulated in Eq.~(\ref{eq:multi-score}),~(\ref{eq:flip-score}), and~(\ref{eq:drop-score}), our method MID only makes changes in selecting top $\lceil k \times n \rceil  $ nodes and is not affected by the input order. Therefore, our proposed MID would not break the graph permutation equivariant property of backbone models, such as TopKPool~\cite{graph-u-net}, SAGPool~\cite{sagpool}, GSAPool~\cite{gsapool}, and ASAP~\cite{asap}.

\subsubsection{Expressiveness Power}

\begin{figure}[!t] 
\setlength{\abovecaptionskip}{-0.1cm}   
\begin{center}
\includegraphics[width=1.0\linewidth]{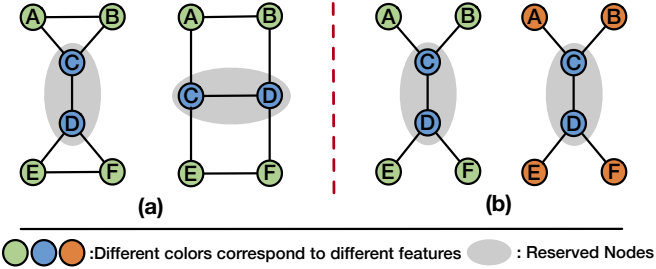}
\end{center}
\caption{Several example graphs which show that MID increases the expressiveness of base node drop pooling models in different cases.}
\label{fig:proof-new}
\end{figure}

\begin{remark}
	MID increases the expressiveness of base node drop pooling models.
\end{remark}

Motivated by the analyses of~\cite{dropgnn, gmt}, we begin with several example graphs, which are not distinguishable in the regular node drop pooling models but can be separated by pooling models combined with MID. 

\textbf{Example 1.} Fig.~\ref{fig:proof-new} (a) shows a fundamental example of two graphs that cannot be distinguished by a Weisfeiler-Lehman (WL) test~\cite{wl-test}, and obviously cannot be distinguished by a node drop pooling model since the set of the reserved nodes of two graphs ($\mathcal{S}_{\text{pool-1}}$, $\mathcal{S}_{\text{pool-2}}$ ) predicted by pooling models would be the same. However, it is not hard to distinguish the two graphs with the dropscore operation of MID. On the condition that $\mathcal{S}_{\text{pool-1}} \cap \mathcal{S}_{\text{dropscore}} \neq \mathcal{S}_{\text{pool-2}} \cap \mathcal{S}_{\text{dropscore}} $, where $\mathcal{S}_{\text{dropscore}}$ is the set of nodes  (correspond to scores) dropped by the dropscore operation, our MID can distinguish the two graphs. For example, node $C$ is dropped in the left graph of Fig.~\ref{fig:proof-new} (a), while node $A$ is dropped in the right graph.

\textbf{Example 2.} Fig.~\ref{fig:proof-new} (b) shows an example of two graphs that can be separated by a regular GCN but cannot be distinguished by a node drop pooling model since the set of the reserved nodes of two graphs predicted by pooling models would be the same. Under this sense, we show that MID could also increase the expressive power of base pooling models. As long as the set of nodes $\mathcal{S}_{\text{dropscore}}$ (correspond to scores) dropped by the dropscore operation in one graph contains the reserved nodes predicted by the base node drop pooling models, $\mathcal{S}_{\text{pool-1}} \cap \mathcal{S}_{\text{dropscore}} \neq \emptyset $, our MID can distinguish the two graphs. For example, even if two graphs both drop the same node $C$ by the dropscore operation, the pooling models would generate two distinguishable embeddings for two graphs.

\subsection{Summary of MID}

\begin{figure}[!t] 
\setlength{\abovecaptionskip}{-0.1cm}   
\begin{center}
\includegraphics[width=1.0\linewidth]{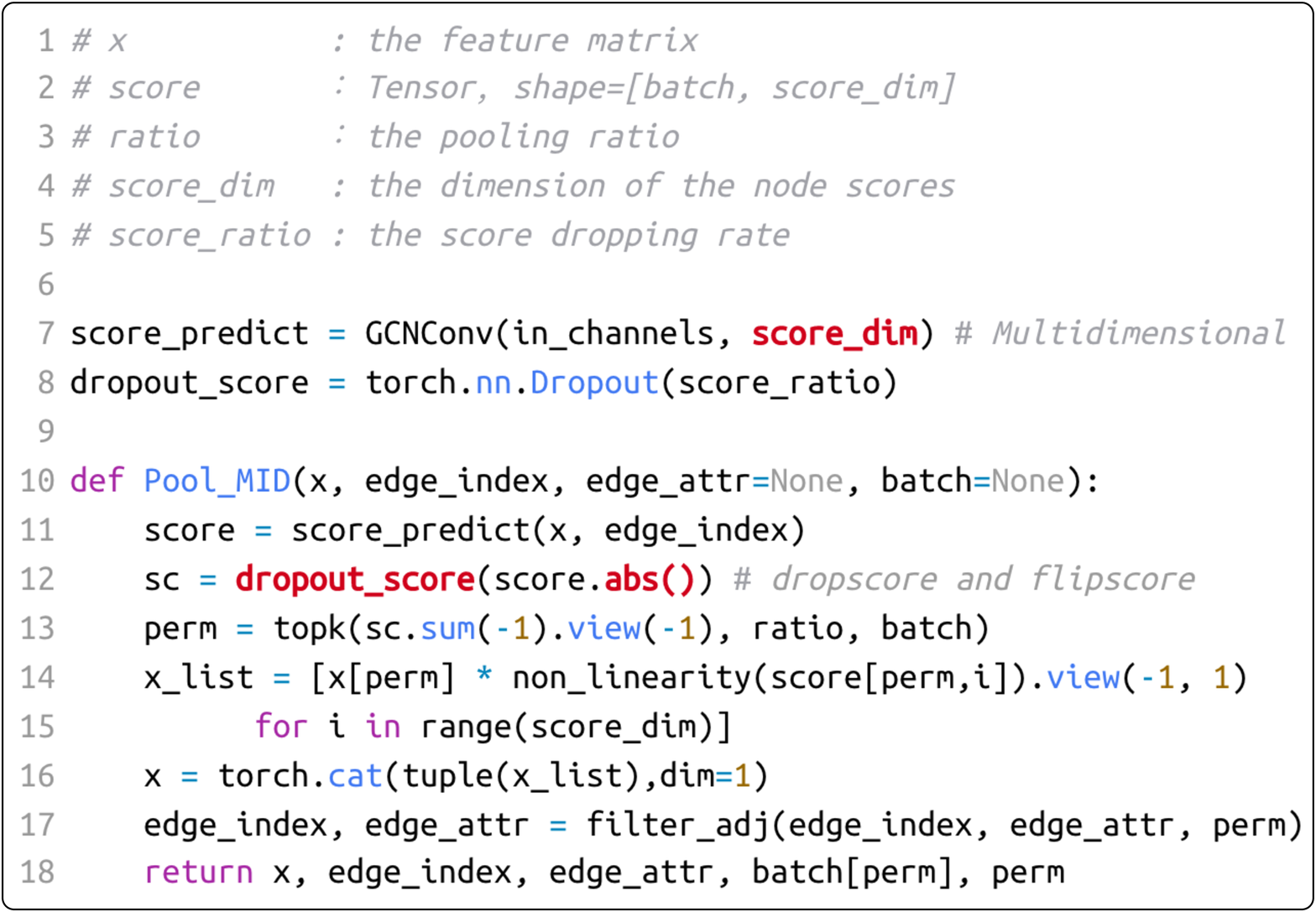}
\end{center}
\caption{PyTorch implementation of MID.}
\label{fig:code-mid}
\end{figure}

Fig.~\ref{fig:model} illustrates the architecture of MID. Specifically, we first predict a multidimensional score for each node by Eq.~(\ref{eq:multi-score}) so that the significance of nodes can be evaluated from multi-views. Then, we apply two operations, flipscore and dropscore, to the generated score matrix with the aim of capturing the node-feature and graph-structure diversities by Eq.~(\ref{eq:flip-score}) and~(\ref{eq:drop-score}), respectively. We discuss and experimentally validate the above three operations that, when appropriately combined with a baseline node drop pooling model such as SAGPool, can encourage the model to maintain the diversity of graph. We also give an in-depth analysis regarding the propositions of the trapped scores and the graph permutation equivariance, and then demonstrate that our proposed MID could increase the expressive power of base node drop pooling models.  

Finally, we present the code of MID in Fig.~\ref{fig:code-mid} with the SAGPool as a backbone model. Due to the simple form, MID can be implemented in PyTorch~\cite{pytorch} with less than 10 lines of code. And the core of MID is highlighted in {\color[HTML]{FE0000} red}.

\section{Experiment}
\label{experiment}

We validate our method, mainly focusing on the task of graph classification, and the results illustrate the advantages of MID in terms of performance, robustness, generalization, and efficiency.

\begin{itemize}
  \item  \textbf{Performance.} MID achieves improvements over all 4 node drop pooling models (averaged across datasets):  3.96\% (SAGPool), 3.97\% (TopKPool), 0.94\% (ASAP), and 3.46\% (GSAPool).
  \item \textbf{Robustness.} As the perturbation rate increases,  TopKPool with MID shows a weak accuracy-declining trend, while the decline of base TopKPool is quite sharp.
  \item \textbf{Generalization.} The accuracy of TopKPool model with MID on COLORS-3 dataset does not degrade significantly when the model is generalized to graphs with larger sizes during the test.
  \item \textbf{Efficiency.} Our MID is highly efficient in terms of time and memory compared with pooling baselines.
\end{itemize}

To fully exploit the expressive power of our method, we also conduct our method  on graph reconstruction task.

\subsection{Graph Classification}

\begin{table}[!t]
\renewcommand\arraystretch{1.2} 
\centering
\caption{Statistics and properties of benchmark datasets}
\label{tab:dataset}
\begin{threeparttable}[b]
\resizebox{0.48\textwidth}{!}{%
\begin{tabular}{@{}lllccc@{}}
\toprule
                              &  \textbf{ \# datasets}            & \textbf{ \# graphs} & \textbf{\# classes} & \textbf{Avg \# nodes}         & \textbf{Avg \# edges}         \\ \midrule
\multirow{13}{*}{\textbf{TUDataset}\tnote{1}} & D\&D           & 1,178     & 2          & 284.32               & 715.66               \\
                              & PROTEINS     & 1,113     & 2          & 39.06                & 72.82                \\
                              & NCI1         & 4,110     & 2          & 29.87                & 32.30                \\
                              & MUTAG        & 188       & 2          & 17.93                & 19.79                \\
                              & PTC-MR       & 344       & 2          & 14.30                & 14.69                \\
                              & NCI109       & 4,127     & 2          & 29.68                & 32.13                \\
                              & ENZYMES      & 600       & 6          & 32.63                & 124.20               \\
                              & MUTAGENICITY & 4,337     & 2          & 30.32                & 30.77                \\
                              & FRANKENSTEIN & 4,337     & 2          & 16.90                & 17.88                \\
                              & REDDIT-BINARY     & 2,000     & 2          & 429.63               & 497.75               \\
                              & IMDB-BINARY       & 1,000     & 2          & 19.77                & 96.53                \\
                              & IMDB-MULTI       & 1,500     & 3          & 13.00                & 65.94                \\
                              & COLLAB       & 5,000     & 3          & 74.49                & 2457.78              \\ \midrule
\multirow{4}{*}{\textbf{OGB}\tnote{2}}  & HIV          & 41,127    & 2          & 25.51                & 27.52                      \\
                              & TOX21        & 7,831     & 12         & 18.57                & 19.3                 \\
                              & TOXCAST      & 8,576     & 617        & 18.78                & 19.3                 \\
                              & BBBP         & 2,039     & 2          & 24.06                & 26.0                 \\ \midrule
\multirow{2}{*}{\textbf{Synthetic}\tnote{3}}  & COLORS-3          & 5,500   & 11          & 61.31                & 91.03                      \\
                              & TRIANGLES        & 45,000     & 10         & 20.85                & 32.74                 \\
                             
\bottomrule
\end{tabular}%
}\begin{tablenotes}
  \scriptsize  
  \item[1] ~\url{https://chrsmrrs.github.io/datasets/docs/datasets/} 
  \item[2] ~\url{https://ogb.stanford.edu/docs/graphprop/} 
  \item[3] ~\url{https://github.com/bknyaz/graph_attention_pool/tree/master/data} 
  \end{tablenotes}
  \end{threeparttable}
\end{table}

\begin{table*}[!t]
\setlength{\belowcaptionskip}{0.3cm}
\renewcommand\arraystretch{1.1} 
\setlength\tabcolsep{2pt} 
\centering
\caption{MID performance across four backbone models and seventeen datasets in graph classification task. The reported results are mean and standard deviation over 100 different runs. \textbf{Bold:}  best performance per backbone model and dataset. {\color[HTML]{FE0000} Red:} the best performance per dataset. {\color[HTML]{3166FF} Blue:} the second best performance per dataset. Hyphen(-) denotes out-of-resources,  i.e., either we could not fit a batch size of 16 graphs on an Nvidia V100 GPU or it took more than 7 days to complete training. The \textbf{Avg. rank} column indicates the average ranking of the methods across all datasets.}
\label{tab:graph_classification}

\resizebox{\textwidth}{!}{%
\begin{tabular}{@{}lccccccccc@{}}
\toprule

\textbf{} &
  \multicolumn{9}{c}{\textbf{\normalsize  Biochemical Domain in TU Datasets (9)}}
   \\ \cmidrule(lr){2-10}
 &
  \textbf{D\&D} &
  \textbf{PROTEINS} &
  \textbf{NCI1} &
  \textbf{MUTAG} &
  \textbf{PTC-MR} &
  \textbf{NCI109} &
  \textbf{ENZYMES} &
  \textbf{MUTAGEN.} & 
  \textbf{FRAN.}
   \\ \midrule
Set2set &
  $72.65_{\pm 0.47}$ &
  $73.14_{\pm 0.97}$ &
  $71.70_{\pm 0.73}$ &
  $70.50_{\pm 1.99}$ &
  $54.53_{\pm 1.69}$ &
  $69.78_{\pm 0.43}$ &
  $42.92_{\pm 2.05}$ &
  $79.86_{\pm 0.43}$ &
  $61.79_{\pm 0.23}$
   \\
SortPool &
  $77.50_{\pm 0.50}$&
  $74.16_{\pm 0.53}$ &
  $72.88_{\pm 0.93}$ &
  $70.56_{\pm 2.73}$ &
  $52.62_{\pm 2.11}$ &
  $71.77_{\pm 0.67}$ &
  $36.17_{\pm 2.58}$ &
  $77.03_{\pm 0.51}$ &
  $62.42_{\pm 0.57}$  \\
DiffPool &
  $67.95_{\pm 2.44}$ &
  $72.86_{\pm 1.00}$ &
  ${\color[HTML]{3166FF} \textbf{77.04}}_{\pm 0.73}$ &
  $82.50_{\pm 2.54}$ &
  $55.26_{\pm 3.84}$&
  $75.38_{\pm 0.66}$ &
  ${\color[HTML]{FE0000} \textbf{51.27}}_{\pm 2.89}$ &
  $79.80_{\pm 0.24}$ &
  $ {\color[HTML]{3166FF} \textbf{63.95}}_{\pm 0.81}$ \\
EdgePool &
  $76.97_{\pm 0.44}$ &
  $74.82_{\pm 0.56}$ &
  $76.53_{\pm 0.50}$ &
  $73.00_{\pm 0.87}$ &
  ${\color[HTML]{FE0000} \textbf{58.09}}_{\pm 2.04}$&
  $75.58_{\pm 0.50}$ &
  $31.18_{\pm 1.99}$ &
  ${\color[HTML]{3166FF} \textbf{80.78}}_{\pm 0.17}$&
  $58.09_{\pm 2.04}$  \\
MinCutPool &
  $74.69_{\pm 0.52}$ &
  $74.28_{\pm 0.76}$ &
  $71.15_{\pm 0.98}$ &
  ${\color[HTML]{FE0000} \textbf{85.17}}_{\pm 1.41}$ &
  $54.68_{\pm 2.45}$ &
  $71.68_{\pm 0.46}$ &
  $25.33_{\pm 1.47}$ &
  $75.39_{\pm 0.39}$ &
  $62.32_{\pm 0.55}$  \\
 HaarPool &
  -- &
  -- &
  -- &
  $74.89_{\pm 0.69}$ &
  $55.97_{\pm 1.26}$ &
  -- &
  $20.57_{\pm 1.05}$ &
  -- &
  --  \\
MemPool &
  $72.96_{\pm 0.84}$ &
  $71.99_{\pm 0.65}$ &
  $67.65_{\pm 0.67}$ &
  $69.17_{\pm 2.04}$ &
  $53.26_{\pm 2.67}$ &
  $65.50_{\pm 1.16}$ &
  ${\color[HTML]{3166FF} \textbf{44.73}}_{\pm 2.21}$&
  $74.96_{\pm 0.22}$ &
  $62.05_{\pm 0.45}$  \\
GMT &
 ${\color[HTML]{FE0000} \textbf{78.48}}_{\pm 0.48}$&
 ${\color[HTML]{FE0000} \textbf{75.08}}_{\pm 0.85}$ &
  $76.23_{\pm 0.49}$ &
  ${\color[HTML]{3166FF} \textbf{83.04}}_{\pm 1.01}$&
  $55.41_{\pm 1.30}$ &
  $74.49_{\pm 0.52}$ &
  $37.38_{\pm 1.52}$ &
  $79.94_{\pm 0.35}$ &
  $63.15_{\pm 0.25}$  \\ \midrule
SAGPool &
  $76.15_{\pm 0.73}$ &
  $73.42_{\pm 0.93}$ &
  $70.89_{\pm 0.80}$ &
  $67.67_{\pm 3.56}$ &
  $54.38_{\pm 1.96}$ &
  $70.26_{\pm 0.99}$ &
  $36.30_{\pm 2.51}$ &
  $74.20_{\pm 0.75}$ &
  $60.62_{\pm 0.66}$  \\
\textbf{SAGPool + MID} &
  $\textbf{76.97}_{\pm 0.72}$ &
  $\textbf{74.52}_{\pm 0.52}$ &
  $\textbf{74.08}_{\pm 0.40}$ &
  $\textbf{72.72}_{\pm 1.96}$ &
  $\textbf{56.24}_{\pm 1.22}$ &
  $\textbf{75.21}_{\pm 0.66}$ &
  $\textbf{39.88}_{\pm 1.93}$ &
  $\textbf{79.75}_{\pm 0.17}$ &
  $\textbf{61.33}_{\pm 0.28}$  \\ \midrule
TopKPool &
  $72.12_{\pm 1.22}$ &
  $72.72_{\pm 1.16}$ &
  $73.82_{\pm 0.65}$ &
  $77.06_{\pm 2.25}$ &
  $55.59_{\pm 2.43}$ &
  $72.94_{\pm 0.67}$ &
  $29.77_{\pm 1.74}$ &
  $76.51_{\pm 1.13}$ &
  $64.63_{\pm 0.67}$  \\
\textbf{TopKPool + MID} &
  $\textbf{74.65}_{\pm 1.68}$ &
  $\textbf{73.07}_{\pm 0.80}$ &
  ${\color[HTML]{FE0000} \textbf{78.50}}_{\pm 0.50}$ &
  $\textbf{80.38}_{\pm 2.26}$ &
  ${\color[HTML]{3166FF} \textbf{56.76}}_{\pm 1.76}$ &
  ${\color[HTML]{FE0000} \textbf{77.55}}_{\pm 0.49}$ &
  $\textbf{30.37}_{\pm 1.94}$ &
  ${\color[HTML]{FE0000} \textbf{81.39}}_{\pm 0.35}$ &
  $ {\color[HTML]{FE0000} \textbf{66.30}}_{\pm 0.55}$  \\ \midrule
ASAP &
  $75.91_{\pm 1.01}$ &
  $71.25_{\pm 0.79}$ &
  $73.86_{\pm 0.74}$ &
  $79.33_{\pm 4.02}$ &
  $55.68_{\pm 1.45}$ &
  $73.15_{\pm 0.70}$ &
  $20.10_{\pm 1.13}$ &
  $77.31_{\pm 0.62}$ &
  $60.57_{\pm 0.62}$  \\
\textbf{ASAP + MID} &
 ${\color[HTML]{3166FF} \textbf{77.57}}_{\pm 0.79}$ &
  $\textbf{72.14}_{\pm 0.67}$ &
  $\textbf{75.30}_{\pm 0.43}$ &
  $\textbf{82.33}_{\pm 3.40}$ &
  ${\color[HTML]{3166FF} \textbf{56.76}}_{\pm 1.88}$ &
  ${\color[HTML]{3166FF} \textbf{75.60}}_{\pm 0.63}$ &
  $\textbf{21.67}_{\pm 1.25}$ &
  $\textbf{78.94}_{\pm 0.50}$ &
  $\textbf{61.55}_{\pm 0.95}$  \\ \midrule
GSAPool &
  $75.91_{\pm 0.77}$ &
  $73.64_{\pm 0.89}$ &
  $71.33_{\pm 0.92}$ &
  $68.83_{\pm 1.54}$ &
  $53.59_{\pm 2.69}$ &
  $70.01_{\pm 1.60}$ &
  $34.93 _{\pm 2.04}$ &
  $76.56_{\pm 0.77}$ &
  $60.41_{\pm 0.53}$  \\
\textbf{GSAPool + MID} &
  $\textbf{76.70}_{\pm 0.56}$ &
  ${\color[HTML]{3166FF} \textbf{75.04}}_{\pm 0.42}$ &
  $\textbf{75.72}_{\pm 0.92}$ &
  $\textbf{71.72}_{\pm 1.04}$ &
  $\textbf{55.47}_{\pm 2.25}$ &
  $\textbf{75.34}_{\pm 0.37}$ &
  $\textbf{42.48}_{\pm 2.52}$ &
  $\textbf{79.88}_{\pm 0.20}$ &
  $\textbf{61.26}_{\pm 0.37}$  \\ \midrule
\textbf{} &
   &
   &
   &
   &
   &
   &
   &
   &
  \\ \midrule

\textbf{} &
  \multicolumn{4}{c}{\normalsize  \textbf{Social Domain in TU Datasets (4)}} &
  \multicolumn{4}{c}{\normalsize  \textbf{OGB Datasets (4)}} \\ \cmidrule(r){2-5}  \cmidrule(r){6-9} 
 &
  \textbf{IMDB-B} &
  \textbf{IMDB-M} &
  \textbf{REDDIT-B} &
  \textbf{COLLAB} &
  \textbf{HIV} &
  \textbf{BBPB} &
  \textbf{TOX21} &
  \textbf{TOXCAST} &
   \multirow{-2}{*}{\textbf{Avg. rank}} \\ \midrule
Set2set &
  $73.10_{\pm 0.48}$ &
  $50.15_{\pm 0.58}$ &
  $ {\color[HTML]{3166FF} \textbf{90.03}}_{\pm 0.45}$ &
  $79.88_{\pm 0.50}$ &
  $73.42_{\pm 2.34}$ &
  $64.43_{\pm 2.16}$ &
  $73.42_{\pm 0.67}$ &
  $59.76_{\pm 0.65}$ & 
  7.9 \\
SortPool &
  $72.49_{\pm 0.78}$ &
  $49.62_{\pm 0.36}$ &
  $87.00_{\pm 1.03}$ &
  $80.21_{\pm 0.35}$ &
  $71.88_{\pm 1.83}$ &
  $64.33_{\pm 3.10}$ &
  $68.90_{\pm 0.78}$ &
  $59.28_{\pm 0.99}$ &
  9.6 \\
DiffPool &
  $72.99_{\pm 0.65}$  &
  $ {\color[HTML]{3166FF} \textbf{51.03}}_{\pm 0.48}$  &
  -- &
  $79.24_{\pm 0.57}$  &
  $75.05_{\pm 1.71}$ &
  $64.77_{\pm 2.43}$ &
  $75.82_{\pm 0.69}$ &
  $ {\color[HTML]{FE0000} \textbf{65.79}}_{\pm 0.87}$ &
  5.6 \\
EdgePool &
  $72.13_{\pm 0.72}$ &
  $ {\color[HTML]{FE0000} \textbf{51.05}}_{\pm 0.53}$ &
  $89.12_{\pm 1.22}$ & 
  $ {\color[HTML]{FE0000} \textbf{81.22}}_{\pm 0.94}$ &
  $72.15_{\pm 1.56}$ &
  $ {\color[HTML]{FE0000} \textbf{68.56}}_{\pm 1.43}$ &
  $74.54_{\pm 0.79}$ &
  $62.57_{\pm 1.36}$ &
  5.1 \\
MinCutPool &
  $73.05_{\pm 0.80}$ &
  $50.22_{\pm 0.81}$ &
  $86.69_{\pm 0.48}$ &
  $78.78_{\pm 0.61}$ &
  $73.91_{\pm 1.10}$ &
  $66.47_{\pm 1.90}$ &
  $ {\color[HTML]{FE0000} \textbf{78.78}}_{\pm 0.61}$ &
  $63.66_{\pm 1.56}$ &
  7.9 \\
HaarPool &
   $ {\color[HTML]{FE0000} \textbf{73.46}}_{\pm 0.55}$ &
  $50.37_{\pm 0.55}$ &
  -- &
  -- &
  -- &
  -- &
  -- &
  -- &
  7.6 \\
MemPool &
  $71.20_{\pm 0.82}$ &
  $49.91_{\pm 0.76}$ &
  --  &
  --  &
  $73.75_{\pm 1.90}$&
  $66.47_{\pm 1.90}$ &
  $72.05_{\pm 0.93}$ &
  $61.85_{\pm 0.36}$ &
  10.9 \\
GMT &
  $73.10_{\pm 0.44}$ &
  $50.50_{\pm 0.54}$ &
  $88.88_{\pm 0.44}$ &
  $ {\color[HTML]{3166FF} \textbf{80.74}}_{\pm 0.54}$ &
  $ {\color[HTML]{FE0000} \textbf{76.41}}_{\pm 2.32}$ &
  $66.88_{\pm 1.59}$ &
  $ {\color[HTML]{3166FF} \textbf{76.56}}_{\pm 0.90}$ &
  $ {\color[HTML]{3166FF} \textbf{64.53}}_{\pm 0.92}$ &
  3.5 \\ \midrule
SAGPool &
  $71.87_{\pm 0.59}$ &
  $50.42_{\pm 0.45}$ &
  $87.42_{\pm 0.62}$ &
  $79.07_{\pm 0.28}$ &
  $70.19_{\pm 3.66}$ &
  $64.29_{\pm 2.96}$ &
  $69.39_{\pm 1.88}$ &
  $59.09_{\pm 1.38}$ &
  11.4\\
\textbf{SAGPool + MID} &
  $\textbf{73.08}_{\pm 0.30}$ &
  $ {\color[HTML]{FE0000} \textbf{51.05}}_{\pm 0.69}$ &
  $ {\color[HTML]{FE0000} \textbf{91.62}}_{\pm 0.30}$ &
  $\textbf{79.93}_{\pm 0.65}$ &
  $\textbf{74.51}_{\pm 1.31}$ &
  $\textbf{66.54}_{\pm 2.33}$ &
  $\textbf{73.06}_{\pm 0.66}$ &
  $\textbf{60.23}_{\pm 0.64}$ &
  \textbf{5.5} \\ \midrule
TopKPool &
  $71.47_{\pm 0.71}$ &
  $49.55_{\pm 0.58}$ &
  $85.37_{\pm 1.04}$ &
  $77.45_{\pm 0.56}$ &
  $71.24_{\pm 2.97}$ &
  $65.93_{\pm 2.60}$ &
  $68.69_{\pm 2.02}$ &
  $58.63_{\pm 1.56}$ &
  10.9\\
\textbf{TopKPool + MID} &
  $\textbf{72.55}_{\pm 0.73}$ &
  $\textbf{50.38}_{\pm 0.51}$ &
  $83.20_{\pm 0.96}$ &
  $\textbf{78.86}_{\pm 0.52}$ &
  $ {\color[HTML]{3166FF} \textbf{75.11}}_{\pm 2.42}$ &
  $\textbf{66.64}_{\pm 2.33}$ &
  $\textbf{71.24}_{\pm 2.14}$ &
  $\textbf{60.50}_{\pm 1.47}$ &
  \textbf{6.1} \\ \midrule
ASAP &
  $72.25_{\pm 0.83}$ &
  $48.55_{\pm 0.64}$ &
  -- &
  -- &
  $71.60_{\pm 1.71}$ &
  $61.93_{\pm 3.18}$ &
  $70.00_{\pm 1.50}$ &
  $60.32_{\pm 1.34}$ &
  10.9 \\
\textbf{ASAP + MID} &
  $ {\color[HTML]{3166FF} \textbf{73.12}}_{\pm 0.56}$ &
  $\textbf{49.47}_{\pm 0.48}$ &
  -- &
  -- &
  $\textbf{72.50}_{\pm 2.16}$ &
  $\textbf{64.03}_{\pm 1.86}$ &
  $\textbf{71.04}_{\pm 0.92}$ &
  $\textbf{61.04}_{\pm 0.42}$ &
  \textbf{7.9} \\ \midrule
GSAPool &
  $72.41_{\pm 0.57}$ &
  $50.72_{\pm 0.57}$ &
  $87.46_{\pm 0.77}$ &
  $78.97_{\pm 0.33}$ &
  $71.47_{\pm 2.43}$ &
  $64.49_{\pm 3.31}$ &
  $69.18_{\pm 2.05}$ &
  $59.60_{\pm 1.17}$ &
  10.6\\
\textbf{GSAPool + MID} &
  $\textbf{73.06}_{\pm 0.20}$ &
  $\textbf{50.93}_{\pm 0.20}$ &
  $\textbf{88.88}_{\pm 0.42}$ &
  $\textbf{79.44}_{\pm 0.30}$ &
  $\textbf{74.49}_{\pm 1.42}$ &
  ${\color[HTML]{3166FF} \textbf{67.27}}_{\pm 1.86}$ &
  $\textbf{72.61}_{\pm 1.14}$ &
  $\textbf{61.90}_{\pm 0.80}$ &
  \textbf{5.6} \\ \bottomrule
\end{tabular}%
}
\end{table*}

\subsubsection{Experimental Settings}

\indent \indent \textbf{Datasets.} We choose 13 datasets from TU datasets~\cite{tu-dataset}, including 9 datasets from the Biochemical domain and 4 datasets from the Social domain. We also select 4 publicly available and relatively large datasets from the Open Graph Benchmark (OGB) datasets~\cite{ogb-dataset}.  The above seventeen real-world datasets vary in content domains and dataset sizes. The dataset statistics are summarized in Table~\ref{tab:dataset}.

\textbf{Models.}  
Four representative node drop graph pooling  methods are selected as backbone models, including \textbf{TopKPool}~\cite{graph-u-net}, \textbf{SAGPool}~\cite{sagpool}, \textbf{ASAP}~\cite{asap}, and \textbf{GSAPool}~\cite{gsapool}. Our MID can be applied to these methods to further improve their performance. 
Besides, we further present eight graph pooling methods as baseline models for comparision, including \textbf{Set2set}~\cite{set2set}, \textbf{SortPool}~\cite{sortpool}, \textbf{DiffPool}~\cite{diffpool}, \textbf{EdgePool}~\cite{edgepool}, \textbf{MinCutPool}~\cite{mincut},  \textbf{HaarPool}~\cite{haar-pooling},  \textbf{MemPool}~\cite{mem-pool}, and \textbf{GMT}~\cite{gmt}. The detailed descriptions of these models are provided in the supplemental material.

\textbf{Implementation Details.}  We use the same experimental settings  following~\cite{gin,gmt}. Specifically, we employ Adam~\cite{adam} to optimize parameters and adopt early stopping to control the training epochs based on validation loss with patience set as 50. For a fair comparison, we fix the pooling ratio to 0.5 for TU datasets and to 0.25 for OGB datasets in each pooling layer for all models. In addition,  we follow the parameter settings (except for the pooling ratio) for some comparing models if the settings are provided in the corresponding papers. Otherwise, we tune the model parameters for better performance.  We evaluate the model performance on TU datasets with 10-fold cross validation~\cite{sortpool,gin,gmt}, using accuracy for evaluation, and evaluate the performance on OGB datasets with their original data split settings~\cite{ogb-dataset}, using ROC-AUC for evaluation.   More experimental and hyperparameter details are described in the supplemental material.

\textbf{Source Code.} We use the source codes provided by  authors for HaarPool~\footnote[1]{\url{https://github.com/YuGuangWang/HaarPool}} and GMT~\footnote[2]{\url{https://github.com/JinheonBaek/GMT}} models, and the codes from PyTorch Geometric library~\footnote[3]{\url{https://github.com/pyg-team/pytorch_geometric}}~\cite{pytorch-geometric} for the rest of models.

\textbf{Environments.}  \textbf{1) Software.} All models are implemented with Python 3.7, PyTorch 1.9.0 or above (which further requires CUDA 10.2 or above), and PyTorch-Geometric 1.7.3 or above. \textbf{2) Hardware.} Each experiment was run on a single GPU (NVIDIA V100 with a 16 GB memory size), and experiments were run on the server at any given time.

\subsubsection{Overall Performance}

The accuracy results of all methods summarized in Table~\ref{tab:graph_classification} are averaged over \textbf{100 runs} with random weight initializations (10 different seeds through the 10-fold cross validation). We highlight the best performance  \textbf{in bold} per backbone model and dataset, the best performance {\color[HTML]{FE0000} in red} per dataset, and the second best performance {\color[HTML]{3166FF} in blue} per dataset.  
\begin{figure}[!t] 
\setlength{\abovecaptionskip}{-0.1cm}   
\begin{center}
\includegraphics[width=1.0\linewidth]{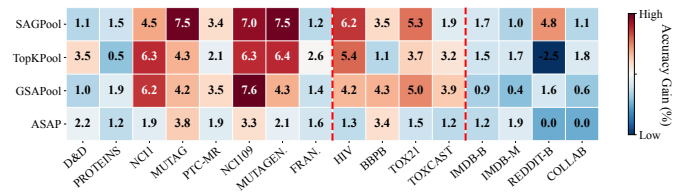}
\end{center}
\label{fig:improvement}
\caption{The improvements by MID across backbone models including SAGPool, TopKPool, GSAPool, and ASAP. We separate the figure into three parts according to the type of datasets by the red dashed lines. From the left to the right: Biochemical domain in TU datasets, Biochemical domain in OGB datasets,  and Social domain in TU datasets.}
\end{figure}

In Table~\ref{tab:graph_classification}, we can observe that MID consistently achieves large-margin enhancement over all backbone models in terms of the average rank across all datasets.  The detailed improvements are more clearly illustrated in Fig.~\ref{fig:improvement}, where we show the improvement achieved by MID on each backbone model and each dataset. From left to right, we split the figure into three parts by the red dashed lines according to the category of datasets. We have the following findings: \textbf{1)} Clearly, MID consistently improves the accuracy of node drop pooling models on all datasets, with a single exception of TopKPool on REDDIT-BINARY. \textbf{2)} MID obtains more significant enhancement on biochemical datasets. Intuitively, this is because the biochemical datasets contain high-quality node features, which are essential for the multidimensional score space and the flipscore operations. \textbf{3)} MID achieves improvements over all 4 node drop pooling models (averaged across datasets):  3.96\% (SAGPool), 3.97\% (TopKPool), 0.94\% (ASAP) and 3.46\% (GSAPool).  MID especially improves SAGPool and TopKPool performance, since several attentions of the two models have been paid to considering the diversity of a graph when they predict the node scores.

\subsubsection{Robustness Analysis}

\begin{figure}[!t] 
\setlength{\abovecaptionskip}{-0.1cm}   
\begin{center}
\includegraphics[width=1.0\linewidth]{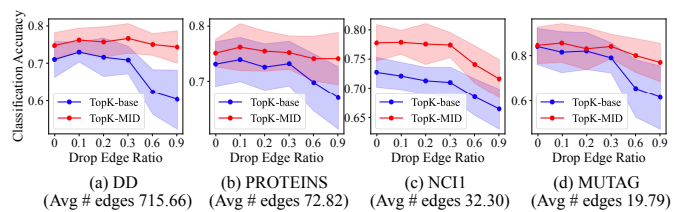}
\end{center}
\caption{The robustness of our method  against edge perturbations using graph classification results of different drop edge ratios.  Solid lines denote the mean, and shaded areas denote the variance. }
\label{fig:robust1}
\end{figure}

We validate the robustness of our method against edge perturbations, which perturb the structure by randomly removing the edges in the pooling layer. In Fig.~\ref{fig:robust1}, we present the classification accuracy of TopKPool model \textbf{w/wo} MID with respect to different perturbation rates on four benchmark datasets. We observe that our method consistently outperforms backbone models across all perturbation rates and all datasets. Especially on D\&D and PROTEINS datasets whose numbers of edges are large, the performance of our method does not degrade significantly with the increase of the perturbation rate compared to the backbone models, which suggests that  MID improves the robustness of backbone pooling models.

\subsubsection{Generalization Improvement}

\begin{table}[!t]
\centering

\setlength\tabcolsep{2pt} 
\caption{Generalization study results.}
\label{tab:generalize}
\begin{threeparttable}[b]
\resizebox{0.48\textwidth}{!}{%
\begin{tabular}{@{}lcccc@{}}
\toprule
               & \multicolumn{2}{c}{\textbf{COLORS-3}}           & \multicolumn{2}{c}{\textbf{TRIANGLES}}       \\ \cmidrule(r){2-3}  \cmidrule(r){4-5} 
               & test-origin      & test-large       & test-origin      & test-large       \\ \midrule
\# nodes train\tnote{1} & 4-25             & 4-25             & 4-25             & 4-25             \\
\# nodes val & 4-25             & 4-25             & 4-25             & 4-25             \\
\# nodes test   & 4-25             & 25-200            & 4-25             & 25-100            \\ \midrule
SAG.-base   & $58.98_{\pm 3.43}$  & $9.68_{\pm 0.60}$  & $44.73_{\pm 0.63}$  & $19.44_{\pm 1.18}$  \\
\textbf{SAG.-MID}   & $\textbf{72.19}_{\pm 1.59}$  & $\textbf{11.24}_{\pm 2.51}$  & $\textbf{59.28}_{\pm 2.05}$  &  $\textbf{22.36}_{\pm 0.97}$               \\  \midrule
TopK.-base   & $70.38_{\pm 5.22}$  & $30.24_{\pm 22.75}$  & $45.00_{\pm 1.06}$  & $17.88_{\pm 1.75}$  \\
\textbf{TopK.-MID}   & $\textbf{99.96}_{\pm 0.03}$  & $\textbf{82.64}_{\pm 15.21}$  & $\textbf{47.30}_{\pm 2.16}$  &  $\textbf{23.85}_{\pm 1.14}$               \\
 \bottomrule
\end{tabular}%
}\begin{tablenotes}
  \scriptsize  
  \item[1] The number of nodes in the training set.  
  \end{tablenotes}
  \end{threeparttable}
\end{table}

Concerned about MID's ability to be generalized to larger and more complex graphs, we test MID on graphs whose sizes are larger than the graph sizes during training following the previous study~\cite{understand-attention}.  We conduct experiments on two synthetic datasets, COLORS-3 and TRIANGLES, introduced by Knyazev et al.~\cite{understand-attention}.  We evaluate MID by applying it to TopKPool and SAGPool models, and strictly follow the experimental settings suggested by~\cite{understand-attention}. The results in Table~\ref{tab:generalize} demonstrate that MID significantly improves the performance across all cases and datasets, and the averaged improvements are about 42.47\%. Furthermore, the accuracy of the TopKPool model combined with MID  does not degrade significantly as the model is generalized to graphs with larger sizes during the test, which demonstrates that MID  significantly improves the generalization capability of pooling models. Moreover, the significant improvements under the \textit{test-origin} case somehow confirms that  MID encourages base pooling models to capture the node-feature diversity since the node features play a significant role in the task of counting colors (COLORS-3 datasets).

\subsubsection{Efficiency Analysis}

\begin{table}[!t]
\centering
\caption{Training time per epoch. The reported values are the average per-epoch training time on all 13 datasets from TU datasets.}
\label{tab:train_time}
\resizebox{0.48\textwidth}{!}{%
\begin{tabular}{@{}llll@{}}
\toprule
\multicolumn{1}{c}{} & \multicolumn{1}{c}{\textbf{ SAGPool}} & \multicolumn{1}{c}{\textbf{ TopKPool}} & \multicolumn{1}{c}{\textbf{GSAPool}} \\ \midrule
Base                 & 0.8358s (1x)                 & 0.7985s (1x)       & 1.0542s (1x)           \\
MID-1               & 0.8487s (1.01x)              & 0.8005s (1.00x)     & 1.0674s (1.01x)          \\
MID-5               & 0.9881s (1.18x)              & 0.9688s (1.21x)    & 1.2251s (1.16x)           \\
MID-9               & 1.2020s (1.44x)              & 1.2052s (1.51x)    & 1.4957s (1.42x)           \\ \bottomrule
\end{tabular}%
}
\end{table}

\indent \indent \textbf{Efficiency Compared with Backbone Models.} As our method is a plug-in one, it is essential to take efficiency into account. The flipscore and dropscore operations introduce negligible overhead, and the efficiency of our method is mainly influenced by the dimension of score $h$. Table~\ref{tab:train_time} reports the average per-epoch training time on all 13 datasets from the TU dataset. We fix the training epochs to 10 with 10 different random seeds and refer to MID with $h$-dimensional score as \textbf{MID$-h$}. It is observed that the additional time consumption keeps relatively low with the increase of the score dimension $h$, which validates that our MID is practically efficient. In addition, Fig.~\ref{fig:hyper} illustrates that increasing $h$ can significantly improve the classification accuracy on biochemical datasets at the cost of its training efficiency. In practice, we can adjust the values of $h$ to balance the trade-off between performance and efficiency.

\begin{figure}[!t] 
\setlength{\abovecaptionskip}{-0.1cm}   
\begin{center}
\includegraphics[width=1.0\linewidth]{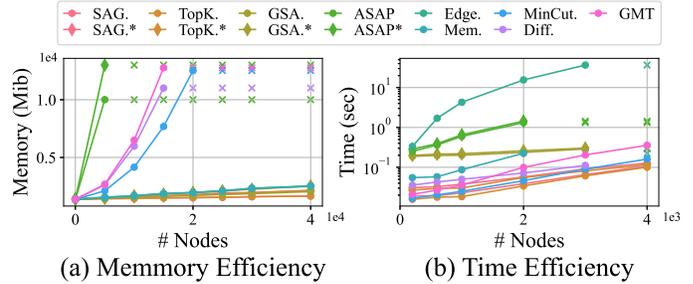}
\end{center}
\caption{Memory and time efficiency of our method compared with baseline models. {\color[HTML]{FF00FF} x} indicates out-of-memory error. SAG. refers to SAGPool and SAG.* refers to SAGPool with MID.} 
\label{fig:efficiency-analysis}
\end{figure}

\textbf{Efficiency Compared with Baseline Models.} While several node clustering methods  achieve decent performances in the graph classification task shown in  Table~\ref{tab:graph_classification}, they suffer from large memory cost due to the usage of a dense soft-assignment matrix. To validate the GPU \textbf{memory efficiency} and \textbf{time efficiency}, we conduct experiments on the Erdos-Renyi graphs~\cite{erdos}. Specifically, we  generate random Erdos-Renyi graphs by varying the number of nodes $n$, where the edge size $m$ is twice  as large as the number of nodes: $m = 2n$, and random 128-dimensional node features.  Fig.~\ref{fig:efficiency-analysis} (a) demonstrates that node drop pooling models are highly efficient in terms of memory compared with memory-heavy pooling baselines such as DiffPool, MinCutPool, and GMT, and our method MID has little impact on backbone models. In addition, Fig.~\ref{fig:efficiency-analysis} (b) demonstrates that node drop pooling models (with or without MID) take less than (or nearly about) a second per epoch even for large graphs, compared to the slowly working models such as MemPool and EdgePool. The above results indicate that node drop pooling models are more practical  and promising on larger real-world datasets.

\subsubsection{Ablation Study}

\begin{table}[!t]
\centering
\setlength\tabcolsep{3pt} 
\caption{Ablation study results. \textbf{Bold:} the best performance per backbone model and dataset.}
\label{tab:ablation}
\resizebox{0.45\textwidth}{!}{%
\begin{tabular}{@{}lcccc@{}}
\toprule
\multirow{2}{*}{} & \multicolumn{2}{c}{\textbf{NCI1}}        & \multicolumn{2}{c}{\textbf{IMDB-BINARY}} \\ \cmidrule(r){2-3}  \cmidrule(r){4-5} 
                       & SAGPool        & TopKPool       & SAGPool        & TopKPool       \\ \midrule
Base                    & 70.89          & 73.82          & 71.87          & 71.47          \\
MID                   & \textbf{76.06} & \textbf{78.78} & \textbf{76.10} & \textbf{73.90} \\ \midrule
w/o flipscore              & 74.76          & 77.22          & 75.90          & 73.80          \\
w/o dropscore             & 74.96          & 78.39          & 74.92          & 72.22          \\
w/o multiscore            & 71.97          & 76.15          & 74.51          & 73.50          \\ \bottomrule

\end{tabular}%
}
\end{table}

We conduct ablation studies to verify that each component of our method contributes to the improvement of performance. For convenience, we name the models without multidimensional score space, flipscore, and dropscore as \textbf{w/o multiscore}, \textbf{w/o flipscore}, and \textbf{w/o dropscore}, respectively. Note that except for the selected component, the rest remain the same as the complete model. We can observe in Table~\ref{tab:ablation} that all variants with some components removed exhibit clear performance drops compared to the complete model, indicating that each component contributes to the improvements. Furthermore, MID without the dropscore operation performs worse on the IMDB-BINARY dataset, demonstrating the significance of the proposed dropscore  technique for datasets without node features.

\subsubsection{Parameter Analysis}

\begin{figure}[!t] 
\setlength{\abovecaptionskip}{-0.1cm}   
\begin{center}
\includegraphics[width=1.0\linewidth]{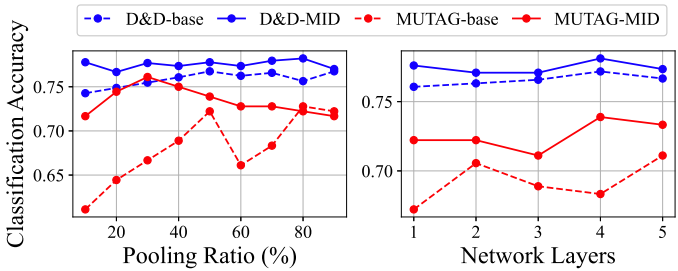}
\end{center}
\caption{Model performance varying with the pooling ratio and the number of model layers. }
\label{fig:analyse}
\end{figure}

\indent \indent \textbf{Inherent Parameter Sensitivity.} We further study how the graph pooling ratio $k$ and the number of pooling layers $L$  would affect the graph classification performance on D\&D and MUTAG datasets with the SAGPool model. As shown in Fig.~\ref{fig:analyse} (a), MID performs better in all cases. In addition,  the accuracy range of  MID is relatively smaller, suggesting that our method selects the nodes that are essential for graph-level representation learning regardless of the pooling ratio. Then, in Fig.~\ref{fig:analyse} (b), we can observe that MID consistently outperforms backbone models when layers go deeper.

\begin{figure}[!t] 
\setlength{\abovecaptionskip}{-0.1cm}   
\begin{center}
\includegraphics[width=1.0\linewidth]{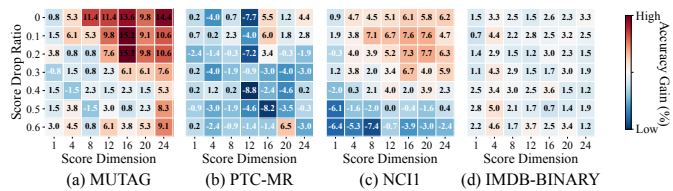}
\end{center}
\caption{Parameter sensitivity of score drop ratio $p_s$ and score dimension $h$ on four datasets.}
\label{fig:hyper}
\end{figure}

\textbf{Introduced Parameter Sensitivity.} We investigate the effects of two new parameters, the dimension of the score $h$ and the score dropping rate $p_s$, introduced by MID with the SAGPool model. In this parameter sensitivity study, $h$ is searched within the range of $\{4,8,12,16,20\}$, while in the experiments, the search space is only $\{5,9\}$. $p_s$ is searched within the range of $\{0.1,0.2,...,0.9\}$, while in all other experiments, the search space is only $\{0.1, 0.2\}$ for all datasets. As shown in Fig.~\ref{fig:hyper}, biochemical datasets (subfigures a, b, and c) prefer a relatively high score dimension and a relatively low score drop ratio, while social datasets (subfigure d) are at the opposite ends. We conjecture that this is because 1) social datasets do not carry node features, the multidimensional score operation would generate $h$ similar scores for each node, and 2) the social datasets generally have more edges, which indicates that they are more robust to the drop operation.

\subsection{Graph Reconstruction}

We further validate MID on the graph reconstruction task, which reveals how much meaningful information is reserved during pooling. 

\begin{figure}[!t] 
\setlength{\abovecaptionskip}{-0.1cm}   
\begin{center}
\includegraphics[width=1.0\linewidth]{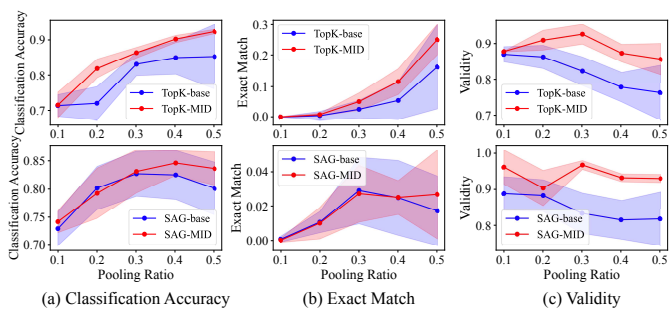}
\end{center}
\caption{Reconstruction results on the ZINC dataset with different pooling ratios. Solid lines denote the mean, and the shaded areas denote the variance.}
\label{fig:reconstruct}
\end{figure}

\textbf{Datasets.} We conduct experiments on a real-world dataset, ZINC~\cite{zinc}, which contains 12K molecular graphs. 

\textbf{Models.} We evaluate MID by applying it to the TopKPool and SAGPool models. Following MinCutPool~\cite{mincut} and GMT~\cite{gmt}, we use two graph convolution layers  before both the pooling operation and the unpooling operation. The pooling operations are TopKPool and SAGPool, and the unpooling operation is proposed in graph U-Net~\cite{graph-u-net}.

\textbf{Experimental Settings.} We perform all experiments 10 times with 10 random seeds, and then report the average results with the standard deviation. We strictly follow the dataset splitting provided by~\cite{benchmarkgnns} and adopt early stopping with the patience set as 50 on the validation loss. We set the maximum number of epochs as 500, the batch size as 128, the hidden dimension as 32, and the pooling ratio of all models as 10\%, 20\%, 30\%, 40\%, and 50\%. The rest of hyperparameters remain the same as those of the above graph classification task.

\textbf{Performance Measure.} We use three metrics suggested by~\cite{gmt, moleculenet} as follows: 1) Classification accuracy refers to the classification accuracy of atom types of all nodes. 2) Exact match indicates the number of reconstructed molecules that are the same as the original molecules. 3) Validity means the number of reconstructed molecules that are chemically valid.

\textbf{Overall Results.} Fig.~\ref{fig:reconstruct} demonstrates that MID significantly enhances the performance of backbone models in terms of the \textbf{validity} metric, suggesting that MID enables backbone models to capture more meaningful nodes in the original molecules. Besides, our method improves the performance of backbone models in the classification accuracy and exact match metrics.  In a nutshell,  MID enhances the expressive power of backbone pooling models in terms of capturing  significant semantic information for the reconstruction of the original graph.

\section{Conclusion}

In this work, we propose an efficient scheme, MID, which improves node drop pooling by exploring the node-feature and graph-structure diversities. Specifically, we first build a multidimensional score space to depict more comprehensive semantic information. Then, two operations on the multidimensional scores, flipscore and dropscore, are devised to reserve the feature diversity by highlighting dissimilar features and to maintain the structure diversity by covering more substructures, respectively. Extensive experiments on seventeen benchmark datasets involving different domains, samples, and graph sizes have verified that MID can generally and consistently improve the performance of current node drop pooling models (\textit{e.g.}, TopKPool, SAGPool, GSAPool, and ASAP). Furthermore, the experimental results indicate that backbone models with MID are much more powerful from the perspective of robustness and generalization.



\ifCLASSOPTIONcaptionsoff
  \newpage
\fi



%
%
%

%

\bibliography{reference.bib}
\bibliographystyle{IEEEtran}

\end{document}